\documentclass{article}

    \PassOptionsToPackage{numbers, compress}{natbib}

\usepackage[final]{neurips_2021}

\usepackage[utf8]{inputenc} 
\usepackage[T1]{fontenc}    
\usepackage{hyperref}       
\usepackage{url}            
\usepackage{booktabs}       
\usepackage{amsfonts, amsmath, amsthm}       
\usepackage{nicefrac}       
\usepackage{microtype}      
\usepackage{xcolor}         
\usepackage{graphicx, subcaption} 
\usepackage{boldline}         
\usepackage{multirow}
\usepackage{siunitx}

\DeclareFontFamily{OMS}{oasy}{\skewchar\font48 }
\DeclareFontShape{OMS}{oasy}{m}{n}{%
         <-5.5> oasy5     <5.5-6.5> oasy6
      <6.5-7.5> oasy7     <7.5-8.5> oasy8
      <8.5-9.5> oasy9     <9.5->  oasy10
      }{}
\DeclareFontShape{OMS}{oasy}{b}{n}{%
       <-6> oabsy5
      <6-8> oabsy7
      <8->  oabsy10
      }{}
\DeclareSymbolFont{oasy}{OMS}{oasy}{m}{n}
\SetSymbolFont{oasy}{bold}{OMS}{oasy}{b}{n}
\DeclareMathSymbol{\smallleftrightarrow}{\mathrel}{oasy}{"24}
\newcommand{\tensor}[1]{\overset{\scriptscriptstyle\smallleftrightarrow}{#1}}
\newcommand{\bR}{\mathbb{R}}
\newcommand{\bg}{\mathbf{g}}
\newcommand{\bh}{\mathbf{h}}
\newcommand{\bI}{\mathbf{I}}
\newcommand{\bJ}{\mathbf{J}}
\newcommand{\bz}{\mathbf{z}}
\newcommand{\bu}{\mathbf{u}}
\newcommand{\bv}{\mathbf{v}}
\newcommand{\bx}{\mathbf{x}}
\newcommand{\cM}{\mathcal{M}}
\newcommand{\cZ}{\mathcal{Z}}
\newcommand{\cU}{\mathcal{U}}
\newcommand{\cV}{\mathcal{V}}
\newcommand{\cR}{\mathcal{R}}
\newcommand{\cX}{\mathcal{X}}

\usepackage{dsfont}

\title{Tractable Density Estimation on Learned Manifolds with Conformal Embedding Flows}

\author{%
  Brendan Leigh Ross \\
  Layer 6 AI \\
  \texttt{brendan@layer6.ai} \And
   Jesse C. Cresswell \\
   Layer 6 AI \\
   \texttt{jesse@layer6.ai} 
}

\begin{document}

\maketitle

\begin{abstract}
Normalizing flows are generative models that provide tractable density estimation via an invertible transformation from a simple base distribution to a complex target distribution. However, this technique cannot directly model data supported on an unknown low-dimensional manifold, a common occurrence in real-world domains such as image data. Recent attempts to remedy this limitation have introduced geometric complications that defeat a central benefit of normalizing flows: exact density estimation. We recover this benefit with Conformal Embedding Flows, a framework for designing flows that learn manifolds with tractable densities. We argue that composing a standard flow with a trainable conformal embedding is the most natural way to model manifold-supported data. To this end, we present a series of conformal building blocks and apply them in experiments with synthetic and real-world data to demonstrate that flows can model manifold-supported distributions without sacrificing tractable likelihoods.
\end{abstract}

\section{Introduction}\label{sec:introduction}

Deep generative modelling is the task of modelling a complex, high-dimensional data distribution from a sample set. Research has encompassed major approaches such as normalizing flows (NFs) \cite{dinh2014, rezende2015}, generative adversarial networks (GANs) \cite{goodfellow2014}, variational autoencoders (VAEs) \cite{kingma2013}, autoregressive models \cite{vanoord2016}, energy-based models \cite{du2019}, score-based models \cite{yang2019}, and diffusion models \cite{ho2020, sohldickenstein2015deep}. NFs in particular describe a distribution by modelling a change-of-variables mapping to a known base density. This approach provides the unique combination of efficient inference, efficient sampling, and exact density estimation, but in practice generated images have not been as detailed or realistic as those of those of other methods \cite{brock2018, child2021, ho2020, karras2019, vahdat2020}.

One limitation of traditional NFs is the use of a base density with the same dimensionality as the data. This stands in contrast to models such as GANs and VAEs, which generate data by sampling from a low-dimensional latent prior and mapping the sample to data space. In many application domains, it is known or commonly assumed that the data of interest lives on a lower-dimensional manifold embedded in the higher-dimensional data space \cite{fefferman2016}. For example, when modelling images, data samples belong to $[0,1]^n$, where $n$ is the number of pixels in each image and each pixel has a brightness in the domain $[0,1]$. However, most points in this data space correspond to meaningless noise, whereas meaningful images of objects lie on a submanifold of dimension $m\ll n$. A traditional NF cannot take advantage of the lower-dimensional nature of realistic images.

There is growing research interest in \textit{injective flows}, which account for unknown manifold structure by incorporating a base density of lower dimensionality than the data space \cite{brehmer2020, cunningham2021, cunningham2020, kothari2021, kumar2020}. Flows with low-dimensional latent spaces could benefit from making better use of fewer parameters, being more memory efficient, and could reveal information about the intrinsic structure of the data. Properties of the data manifold, such as its dimensionality or the semantic meaning of latent directions, can be of interest as well \cite{kingma2013, radford2015}. However, leading injective flow models still suffer from drawbacks including intractable density estimation \cite{brehmer2020} or reliance on stochastic inverses \cite{cunningham2020}.

In this paper we propose Conformal Embedding Flows (CEFs), a class of flows that use \emph{conformal embeddings} to transform from low to high dimensions while maintaining invertibility and an efficiently computable density. We show how conformal embeddings can be used to learn a lower dimensional data manifold, and we combine them with powerful NF architectures for learning densities. The overall CEF paradigm permits efficient density estimation, sampling, and inference. We propose several types of conformal embedding that can be implemented as composable layers of a flow, including three new invertible layers: the orthogonal $k \times k$ convolution, the conditional orthogonal transformation, and the special conformal transformation. Lastly, we demonstrate their efficacy on synthetic and real-world data.

\section{Background}\label{sec:background}
\subsection{Normalizing Flows}\label{sec:background_bijective}
In the traditional setting of a normalizing flow \cite{dinh2014, rezende2015}, an independent and identically distributed sample $\{\mathbf{x}_i\}\subset \mathcal{X}= \mathbb{R}^{n}$ from an unknown ground-truth distribution with density $p_\mathbf{x}^*(\mathbf{x})$ is used to learn an approximate density $p_\mathbf{x}(\mathbf{x})$ via maximum likelihood estimation. The approximate density is modelled using a diffeomorphism $\mathbf{f}: \mathcal{Z}\to \mathcal{X}$ which maps a base density $p_\mathbf{z}(\mathbf{z})$ over the space $\mathcal{Z}=\mathbb{R}^{n}$, typically taken to be a multivariate normal, to $p_\mathbf{x}(\mathbf{x})$ via the change of variables formula
\begin{equation}\label{eq:change-of-variables}
    p_\mathbf{x}(\mathbf{x}) = p_\mathbf{z}\left(\mathbf{f}^{-1}(\mathbf{x})\right)\left\vert \det \mathbf{J}_\mathbf{f}\left(\mathbf{f}^{-1}(\mathbf{x})\right) \right\vert^{-1},
\end{equation}
where $\mathbf{J}_\mathbf{f}(\mathbf{z})$ is the Jacobian matrix of $\mathbf{f}$ at the point $\mathbf{z}$. In geometric terms, the probability mass in an infinitesimal volume $\mathbf{dz}$ of $\mathcal{Z}$ must be preserved in the volume of $\mathcal{X}$ corresponding to the image $\mathbf{f}(\mathbf{dz})$, and the magnitude of the Jacobian determinant is exactly what accounts for changes in the coordinate volume induced by $\mathbf{f}$. By parameterizing classes of diffeomorphisms $\mathbf{f}_{\theta}$, the flow model can be fitted via maximum likelihood on the training data $\{\mathbf{x}_i\}\subset \mathcal{X}$. Altogether, the three following operations must be tractable: sampling with $\mathbf{f}(\mathbf{z})$, inference with $\mathbf{f}^{-1}(\mathbf{x})$, and density estimation with the $\left\vert \det \mathbf{J}_\mathbf{f}\left(\mathbf{z}\right) \right\vert$ factor. To scale the model, one can compose many such layers $\mathbf{f} = \mathbf{f}_k\circ\dots\circ\mathbf{f}_1$, and the $\left\vert \det \mathbf{J}_{\mathbf{f}_i}\left(\mathbf{z}\right) \right\vert$ factors multiply in Eq. \eqref{eq:change-of-variables}.

Generally, there is no unifying way to parameterize an arbitrary bijection satisfying these constraints. Instead, normalizing flow research has progressed by designing new and more expressive component bijections which can be parameterized, learned, and composed. In particular, progress has been made by designing invertible layers whose Jacobian determinants are tractable by construction. A significant theme has been to structure flows to have a triangular Jacobian \cite{dinh2014, dinh2017, kingma2016, rezende2015}. \citet{kingma2018} introduced invertible $1\times 1$ convolution layers for image modelling; these produce block-diagonal Jacobians whose blocks are parameterized in a $PLU$-decomposition, so that the determinant can be computed in $\mathcal{O}(c)$, the number of input channels. See \citet{papamakarios2021} for a thorough survey of normalizing flows.

\subsection{Injective Flows}\label{sec:background_injective}
The requirement that $\mathbf{f}$ be a diffeomorphism fixes the dimensionality of the latent space. In turn, $p_\mathbf{x}(\mathbf{x})$ must have full support over $\mathcal{X}$, which is problematic when the data lies on a submanifold $\cM \subset \mathcal{X}$ with dimension $m < n$. \citet{dai2019} observed that if a probability model with full support is fitted via maximum likelihood to such data, the estimated density can converge towards infinity on $\cM$ while ignoring the true data density $p^*(\mathbf{x})$ entirely. \citet{behrmann2020} point out that analytically invertible neural networks can become numerically non-invertible, especially when the effective dimensionality of data and latents are mismatched. Correctly learning the data manifold along with its density may circumvent these pathologies.

Injective flows seek to learn an explicitly low-dimensional support by reducing the dimensionality of the latent space and modelling the flow as a \textit{smooth embedding}, or an injective function which is diffeomorphic to its image\footnote{Throughout this work we use ``embedding'' in the topological sense: a function which describes how a low-dimensional space can sit inside a high-dimensional space. This is not to be confused with other uses for the term in machine learning, namely a low-dimensional representation of high-dimensional or discrete data.}. This case can be accommodated with a generalized change of variables formula for densities as follows \cite{gemici2016}. 

Let $\mathbf{g}:\mathcal{U}\to\mathcal{X}$ be a smooth embedding from a latent space $\mathcal{U}$ onto the data manifold $\mathcal{M}\subset \mathcal{X}$. That is, $\mathcal{M} = \mathbf{g}(\mathcal{U})$ is the range of $\mathbf{g}$. Accordingly, $\mathbf{g}$ has a left-inverse\footnote{$\dagger$ denotes a left-inverse function, not necessarily the matrix pseudoinverse.} $\mathbf{g}^{\dag}:\mathcal{X} \to \mathcal{U}$ which is smooth on $\cM$ and satisfies $\mathbf{g}^{\dag}(\mathbf{g}(\mathbf{u})) = \mathbf{u} $ for all $\mathbf{u}\in \mathcal{U}$. Suppose $p_\mathbf{u}(\mathbf{u})$ is a density on $\mathcal{U}$ described using coordinates $\mathbf{u}$. The same density can be described in the ambient space $\mathcal{X}$ using $\mathbf{x}$ coordinates by pushing it through $\mathbf{g}$.

The quantity that accounts for changes in the coordinate volume at each point $\mathbf{u}$ is $\sqrt{\det \left[\mathbf{J}_\mathbf{g}^T(\mathbf{u}) \mathbf{J}_\mathbf{g}(\mathbf{u})\right]}$, where the Jacobian $\mathbf{J}_\mathbf{g}$ is now a $n\times m$ matrix \cite{lee2018}. Hence, using the shorthand $\bu = \mathbf{g}^\dagger(\bx)$, the generalized change of variables formula defined for $\mathbf{x}\in \mathcal{M}$ can be written
\begin{equation}\label{eq:generalized-change-of-variables}
    p_\mathbf{x}(\mathbf{x}) = p_\mathbf{u}\left(\bu\right) \left\vert\det \left[ \mathbf{J}_\mathbf{g}^T(\bu) \mathbf{J}_\mathbf{g}(\bu)\right]\right\vert^{-\tfrac{1}{2}}.
\end{equation}

\begin{figure}
    \centering
    \includegraphics[width=1.0\textwidth]{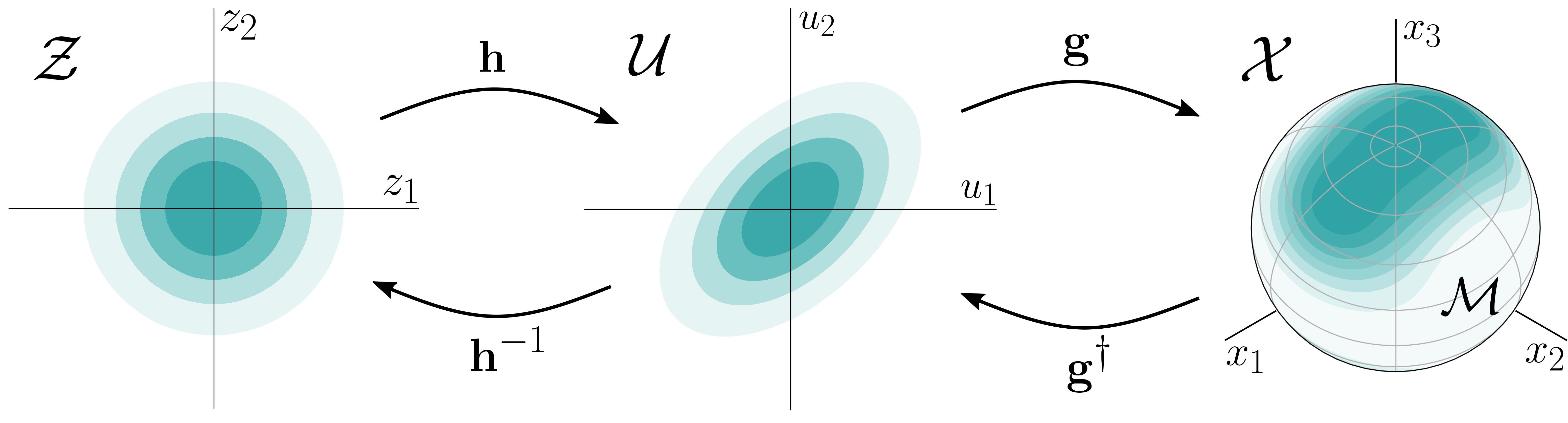}
    \caption{A normalized base density in the $\mathcal{Z}$ space is mapped by a bijective flow $\mathbf{h}$ to a more complicated density in $\mathcal{U}$. The injective component $\mathbf{g}$ maps this density onto a manifold $\mathcal{M}$ in $\mathcal{X}$. For inference, data points from $\mathcal{M}$ follow the reverse path through $\mathbf{g}^\dag$ and $\mathbf{h}^{-1}$ to the latent space $\cZ$ where their densities can be evaluated and combined with the determinant factors in Eq. \eqref{eq:composed-change-of-variables}.} \label{fig:latent-to-data}
\end{figure}

While $\mathbf{g}$ describes how the data manifold $\mathcal{M}$ is embedded in the larger ambient space, the mapping $\mathbf{g}$ alone may be insufficient to represent a normalized base density. As before, it is helpful to introduce a latent space $\mathcal{Z}$ of dimension $m$ along with a diffeomorphism $\mathbf{h}: \mathcal{Z} \to \mathcal{U}$ representing a bijective NF between $\mathcal{Z}$ and $\mathcal{U}$ \cite{brehmer2020}.
Taking the overall injective transformation $\bg \circ \bh$ and applying the chain rule $\mathbf{J}_{\bg \circ \bh} = \mathbf{J}_{\bg}\mathbf{J}_{ \bh}$ simplifies the determinant in Eq. \eqref{eq:generalized-change-of-variables} since the outer Jacobian $\mathbf{J}_\bh$ is square,
\begin{equation}\label{eq:jacobian-simplification}
    \det \left[ \mathbf{J}^T_{\bh}\mathbf{J}^T_{ \bg} \mathbf{J}_{\bg}\mathbf{J}_{ \bh}\right] =  (\det \mathbf{J}_{\bh})^2 \det\left[\mathbf{J}^T_{ \bg} \mathbf{J}_{\bg}\right].
\end{equation}
Finally, writing $\bz = \mathbf{h}^{-1}(\bu)$, the data density is modelled by 
\begin{equation}\label{eq:composed-change-of-variables}
    p_\mathbf{x}(\mathbf{x}) = 
    p_\mathbf{z}\left(\bz\right)
    \left\vert \det \mathbf{J}_{\mathbf{h}}\left(\bz\right) \right\vert^{-1}
    \left\vert\det \left[ \mathbf{J}_\mathbf{g}^T(\bu) \mathbf{J}_\mathbf{g}(\bu)\right]\right\vert^{-\tfrac{1}{2}},
\end{equation}
with the entire process depicted in Fig. \ref{fig:latent-to-data}. 

Generating samples from $p_\mathbf{x}(\mathbf{x})$ is simple; a sample $\mathbf{z}\sim p_\mathbf{z}(\mathbf{z})$ is drawn from the base density and passed through $\mathbf{g}\circ \mathbf{h}$. Inference on a data sample $\mathbf{x}\sim p_\mathbf{x}(\mathbf{x})$ is achieved by passing it through $\mathbf{h}^{-1}\circ \mathbf{g}^\dag$, evaluating the density according to $p_\mathbf{z}(\mathbf{z})$, and computing both determinant factors. 

Notably, the learned density $p_\mathbf{x}(\mathbf{x})$ only has support on a low-dimensional subset $\mathcal{M}$ of $\mathcal{X}$, as per the manifold hypothesis. This formulation leads the learned manifold $\mathcal{M}$ to be diffeomorphic to Euclidean space, which can cause numerical instability when the data's support differs in topology \cite{cornish2020}, but we leave this issue to future work. 

In practice, there will be off-manifold points during training or if $g(\bu)$ cannot perfectly fit the data, in which case the model's log-likelihood will be $-\infty$.  \citet{cunningham2020} remedy this by adding an off-manifold noise term to the model, but inference requires a stochastic inverse, and the model must be optimized using an ELBO-like objective. Other work \cite{brehmer2020, caterini2021, kothari2021} has projected data to the manifold via $\bg \circ \bg^\dag$ prior to computing log-likelihoods and optimized $\bg$ using the reconstruction loss $\mathbb{E}_{\bx\sim p^*_\bx}||\bx - \bg(\bg^\dagger(\bx))||^2$. We prove in App. \ref{app:recon} that minimizing the reconstruction loss brings the learned manifold into alignment with the data manifold.

When computing log-likelihoods, the determinant term $\log\det \left[ \mathbf{J}_\mathbf{g}^T\mathbf{J}_\mathbf{g}\right]$ presents a computational challenge.
\citet{kumar2020} maximize it using an approximate lower bound, while \citet{brehmer2020} and \citet{kothari2021} circumvent its computation altogether by only maximizing the other terms in the log-likelihood. In concurrent work, \citet{caterini2021} optimize injective flows using a stochastic estimate of the log-determinant's gradient. They are also able to optimize $\log\det \left[ \mathbf{J}_\mathbf{g}^T\mathbf{J}_\mathbf{g}\right]$ exactly for smaller datasets, but this procedure involves the explicit construction of $\mathbf{J}_\mathbf{g}$, which would be memory-intensive to scale to larger data such as CelebA. In line with research to build expressive bijective flows where $\det \mathbf{J}_\mathbf{f}$ is tractable, our work focuses on designing and parameterizing injective flows where $\log\det \left[ \mathbf{J}_\mathbf{g}^T\mathbf{J}_\mathbf{g}\right]$ as a whole is efficiently computable. In contrast to past injective flow models, our approach allows for straightforward evaluation and optimization of $\log\det \left[ \mathbf{J}_\mathbf{g}^T\mathbf{J}_\mathbf{g}\right]$ in the same way standard NFs do for $\log \left\vert \det \mathbf{J}_\mathbf{f}\right\vert$. As far as we can find, ours is the first approach to make this task tractable at scale.

\section{Conformal Embedding Flows}\label{sec:method}

In this section we propose Conformal Embedding Flows (CEFs) as a method for learning the low-dimensional manifold $\mathcal{M} \subset \mathcal{X}$ and the probability density of the data on the manifold. 

Modern bijective flow work has produced tractable $\log\vert\det \mathbf{J}_{\mathbf{f}}\vert$ terms by designing layers with triangular Jacobians \cite{dinh2014, dinh2017}. For injective flows, the combination $\mathbf{J}_\mathbf{g}^T\mathbf{J}_\mathbf{g}$ is symmetric, so it is triangular if and only if it is diagonal. In turn, $\mathbf{J}_\mathbf{g}^T\mathbf{J}_\mathbf{g}$ being diagonal is equivalent to $\mathbf{J}_\bg$ having orthogonal columns. While this restriction is feasible for a single layer $\mathbf{g}$, it is not composable. If $\bg_1$ and $\bg_2$ are both smooth embeddings whose Jacobians have orthogonal columns, it need not follow that $\bJ_{\bg_2 \circ \bg_1}$ has orthogonal columns. Additionally, since the Jacobians are not square the determinant in Eq. \eqref{eq:generalized-change-of-variables}, $\det \left[ \mathbf{J}^T_{\bg_1}\mathbf{J}^T_{ \bg_2} \mathbf{J}_{\bg_2}\mathbf{J}_{ \bg_1}\right]$, cannot be factored into a product of individually computable terms as in Eq. \eqref{eq:jacobian-simplification}. To ensure composability we propose enforcing the slightly stricter criterion that each $\mathbf{J}_\mathbf{g}^T\mathbf{J}_\mathbf{g}$ be a scalar multiple of the identity. This is precisely the condition that $\bg$ is a conformal embedding.

Formally, $\bg: \cU \to \cX$ is a \textit{conformal embedding} if it is a smooth embedding whose Jacobian satisfies
\begin{equation}\label{eq:conformal_transformation}
    \mathbf{J}_\mathbf{g}^T(\bu) \mathbf{J}_\mathbf{g}(\bu) = \lambda^2(\bu)\bI_{m}\,,
\end{equation}
 where $\lambda: \cU \to \bR$ is a smooth non-zero scalar function, the \textit{conformal factor} \cite{lee2018}. In other words, $\bJ_\bg$ has orthonormal columns up to a smoothly varying non-zero multiplicative constant. Hence $\bg$ locally preserves angles. 
 
From Eq. \eqref{eq:conformal_transformation} it is clear that conformal embeddings naturally satisfy our requirements as an injective flow. In particular, let $\bg: \cU \to \cX$ be a conformal embedding and $\bh: \cZ \to \cU$ be a standard normalizing flow model. The injective flow model $\bg \circ \bh: \cZ \to \cX$ satisfies
\begin{equation}\label{eq:conformal-density}
    p_\mathbf{x}(\mathbf{x}) = p_\mathbf{z}\left(\bz\right)
    \left\vert \det \mathbf{J}_{\mathbf{h}}\left(\bz\right) \right\vert^{-1}
    \lambda^{-m}(\bu)\,.
\end{equation}

We call $\bg \circ \bh$ a Conformal Embedding Flow. 

CEFs provide a new way to coordinate the training dynamics of the model's manifold and density. It is important to note that not all parameterizations $\bg$ of the learned manifold are equally suited to density estimation \cite{caterini2021}. Prior injective flow models \cite{brehmer2020, kothari2021} have been trained \emph{sequentially} by first optimizing $\bg$ using the reconstruction loss $\mathbb{E}_{\bx\sim p^*_\bx}\Vert\bx - \bg(\bg^\dagger(\bx))\Vert^2$, then training $\bh$ for maximum likelihood with $\bg$ fixed. This runs the risk of initializing the density $p_\mathbf{u}\left(\bu\right)$ in a configuration that is challenging for $\bh$ to learn. \citet{brehmer2020} also alternate training $\bg$ and $\bh$, but this does not prevent $\bg$ from converging to a poor configuration for density estimation. Unlike previous injective flows, CEFs have tractable densities, which allows $\bg$ and $\bh$ to be trained \emph{jointly} by optimizing the loss function 
\begin{equation}\label{eq:mixed-loss-function}
    \mathcal{L} = \mathbb{E}_{\bx\sim p^*_\bx}\left[-\log p_\bx(\bx) + \alpha\Vert\bx - \bg(\bg^\dagger(\bx))\Vert^2\right].
\end{equation}
This mixed loss provides more flexibility in how $\bg$ is learned, and is unique to our model because it is the first for which $\log p_\bx(\bx)$ is tractable.

\subsection{Designing Conformal Embedding Flows}

Having established the model's high-level structure and training objective, it remains for us to design conformal embeddings $\bg$ which are capable of representing complex data manifolds. For $\bg$ to be useful in a CEF we must be able to sample with $\bg(\bu)$, perform inference with $\bg^{\dagger}(\bx)$, and compute the conformal factor $\lambda(\bu)$. In general, there is no unifying way to parameterize the entire family of such conformal embeddings (see App. \ref{app:conformal} for more discussion). As when designing standard bijective flows, we can only identify subfamilies of conformal embeddings which we parameterize and compose to construct an expressive flow. To this end, we work with conformal building blocks $\bg_i: \cU_{i-1} \to \cU_{i}$ (where $\cU_{0} = \cU$ and $\cU_k = \cX$), which we compose to produce the full conformal embedding $\bg$:
\begin{equation}
    \bg = \bg_k \circ \cdots \circ \bg_1\,.
\end{equation}
In turn, $\bg$ is conformal because
\begin{equation}\label{eq:conformal_composition}
    \mathbf{J}_\mathbf{g}^T \mathbf{J}_\mathbf{g} 
    =\left(\mathbf{J}_{\mathbf{g}_1}^T \cdots\mathbf{J}_{\mathbf{g}_k}^T\right)\left(\mathbf{J}_{\mathbf{g}_k} \cdots\mathbf{J}_{\mathbf{g}}\right)
    = \lambda_1^2\cdots \lambda_k^2\bI_{m}\,.
\end{equation}
Our goal in the remainder of this section is to design classes of conformal building blocks which can be parameterized and learned in a CEF.

\subsubsection{Conformal Embeddings from Conformal Mappings}\label{subsec:conf-same-dim}

Consider the special case where the conformal embedding
maps between Euclidean spaces $\cU \subseteq \bR^d$ and $\cV \subseteq \bR^d$ of the same dimension\footnote{We consider conformal mappings between spaces of dimension $d>2$. Conformal mappings in $d=2$ are much less constrained, while the case $d=1$ is trivial since there is no notion of an angle.}. In this special case $\bg_i$ is called a \textit{conformal mapping}.
Liouville's theorem \cite{hartman1958} states that any conformal mapping can be expressed as a composition of translations, orthogonal transformations, scalings, and inversions, which are defined in Table
\ref{tab:trans_rot_scale_sct} (see App. \ref{app:conformal_transformations} for details on conformal mappings). We created conformal embeddings primarily by composing these layers. Zero-padding \cite{brehmer2020} is another conformal embedding, with Jacobian $\bJ_{\bg} = \left( \bI_{m} \ 0\right)^T$, and can be interspersed with conformal mappings to provide changes in dimensionality.

 \begingroup 
\setlength{\tabcolsep}{5pt} 
\renewcommand{\arraystretch}{1.5} 
\begin{table}[ht]
\small
\caption{Conformal Mappings} \label{tab:trans_rot_scale_sct}
\centering
\begin{tabular*}{\textwidth}{ l | l | l | l | l } 
\toprule
 \textsc{Type} & \textsc{Functional Form} & \textsc{Params} & \textsc{Inverse} & $\lambda(\bu)$ \\
\midrule
Translation & $\mathbf{u} \mapsto \mathbf{u} + \mathbf{a}$ & $\mathbf{a}\in \mathbb{R}^d$ & $\mathbf{v} \mapsto \mathbf{v} - \mathbf{a}$& $1$ \\
  \hline
 Orthogonal &  $\mathbf{u} \mapsto \mathbf{Q u}$ & $\mathbf{Q}\in O(d)$ & $\mathbf{v} \mapsto \mathbf{Q}^T\mathbf{v}$ & $1$  \\
  \hline
 Scaling & $\mathbf{u} \mapsto \lambda \mathbf{u}$ & $\lambda \in \mathbb{R}$ & $\mathbf{v} \mapsto \lambda^{-1} \mathbf{v}$& $\lambda$  \\
  \hline
 Inversion & $\mathbf{u} \mapsto \mathbf{u}/\Vert \mathbf{u}\Vert^2$ & & $\mathbf{v} \mapsto \mathbf{v}/\Vert \mathbf{v}\Vert^2$ & $\Vert \mathbf{u}\Vert^{-2}$  \\
  \hline
 SCT & $\mathbf{u} \mapsto \frac{\mathbf{u}-\Vert\mathbf{u}\Vert^2\mathbf{b}}{1- 2\mathbf{b}\cdot\mathbf{u} + \Vert{\mathbf{b}}\Vert^2\Vert{\mathbf{u}}\Vert^2}$ & $\mathbf{b}\in \mathbb{R}^d$ & $\mathbf{v} \mapsto \frac{\mathbf{v}+\Vert\mathbf{v}\Vert^2\mathbf{b}}{1+ 2\mathbf{b}\cdot\mathbf{v} + \Vert{\mathbf{b}}\Vert^2\Vert{\mathbf{v}}\Vert^2}$ & $ 1{-}2\mathbf{b}\cdot\mathbf{u}{+}\Vert \mathbf{b}\Vert^2 \Vert\mathbf{u}\Vert^2$ \\
 \bottomrule
\end{tabular*}
\end{table}
\endgroup

Stacking translation, orthogonal transformation, scaling, and inversion layers is sufficient to learn any conformal mapping in principle. However, the inversion operation is numerically unstable, so we replaced it with the \textit{special conformal transformation} (SCT), a transformation of interest in conformal field theory \cite{francesco2012}. It can be understood as an inversion, followed by a translation by $-\mathbf{b}$, followed by another inversion. In contrast to inversions, SCTs have a continuous parameter and include the identity when this parameter is set to 0.

The main challenge to implementing conformal mappings was writing trainable orthogonal layers. We parameterized orthogonal transformations in two different ways: by using Householder matrices \cite{tomczak2016}, which are cheaply parameterizable and easy to train, and by using GeoTorch, the API provided by \cite{lezcano2019b}, which parameterizes the special orthogonal group by taking the matrix exponential of skew-symmetric matrices. GeoTorch also provides trainable non-square matrices with orthonormal columns, which are conformal embeddings (not conformal mappings) and which we incorporate to change the data's dimensionality.

To scale orthogonal transformations to image data, we propose a new invertible layer: the \textit{orthogonal $k \times k$ convolution}. In the spirit of the invertible $1\times1$ convolutions of \citet{kingma2018}, we note that a $k \times k$ convolution with stride $k$ has a block diagonal Jacobian. The Jacobian is orthogonal if and only if these blocks are orthogonal. It suffices then to convolve the input with a set of filters that together form an orthogonal matrix. Moreover, by modifying these matrices to be non-square with orthonormal columns (in practice, reducing the filter count), we can provide conformal changes in dimension. It is also worth noting that these layers can be inverted efficiently by applying a transposed convolution with the same filter, while a standard invertible $1\times 1$ convolution requires a matrix inversion. This facilitates quick forward and backward passes when optimizing the model's reconstruction loss.

\subsubsection{Piecewise Conformal Embeddings}

To make the embeddings more expressive, the conformality condition on $\bg$ can be relaxed to the point of being conformal \textit{almost everywhere}. Formally, the latent spaces $\cZ$ and $\cU$ are redefined as $\cZ = \{\bz: \bg \text{ is conformal at } \bh(\bz)\}$ and $\cU= \bh(\cZ)$. Then $\bg$ remains a conformal embedding on $\cU$, and as long as $\{\bx: \bg \text{ is nonconformal at }\bg^\dagger(\bx)\}$ also has measure zero, this approach poses no practical problems. Note that the same relaxation is performed implicitly with the diffeomorphism property of standard flows when rectifier nonlinearities are used in coupling layers \cite{dinh2017} and can be justified by generalizing the change of variables formula \cite{koenen2021generalization}.

 \begingroup 
\setlength{\tabcolsep}{7.5pt} 
\renewcommand{\arraystretch}{1.5} 
\begin{table}[ht]
\small
\caption{Piecewise Conformal Embeddings} \label{tab:piecewise-conformal}
\centering
\begin{tabular*}{\textwidth}{ l | l | l | l | l } 
\toprule
 \textsc{Type} & \textsc{Functional Form} & \textsc{Params} & \textsc{Left Inverse} & $\lambda(\bu)$ \\
\midrule
\begin{tabular}{@{}l@{}}Conformal \\ReLu \end{tabular}
 & $\mathbf{u} \mapsto \text{ReLU}
 \begin{bmatrix}
 \mathbf{Qu}\\
 -\mathbf{Qu}
 \end{bmatrix}$ &$\mathbf{Q}\in O(d)$ &
 $\begin{bmatrix}
 \bv_1\\
 \bv_2
 \end{bmatrix} \mapsto \mathbf{Q}^T\left(\bv_1 - \bv_2\right)$ & $1$  \\
 \hline
 \begin{tabular}{@{}l@{}}Conditional \\Orthogonal \end{tabular}
 & $\mathbf{u} \mapsto
 \begin{cases}
 \mathbf{Q}_1 \mathbf{u} & \text{ if $\Vert\mathbf{u}\Vert < 1$}\\
 \mathbf{Q}_2 \mathbf{u} & \text{ if $\Vert\mathbf{u}\Vert \geq 1$}\\
 \end{cases}$ & $\mathbf{Q}_1, \mathbf{Q}_2\in O(d)$ & $\mathbf{v} \mapsto
 \begin{cases}
 \mathbf{Q}_1^T \mathbf{u} & \text{ if $\Vert\mathbf{v}\Vert < 1$}\\
 \mathbf{Q}_2^T \mathbf{u} & \text{ if $\Vert\mathbf{v}\Vert \geq 1$}\\
 \end{cases}$ & $1$  \\
 \bottomrule
\end{tabular*}
\end{table}
\endgroup

We considered the two piecewise conformal embeddings defined in Table \ref{tab:piecewise-conformal}. Due to the success of ReLU in standard deep neural networks \cite{nair2010}, we try a ReLU-like layer that is piecewise conformal. \textit{conformal ReLU} is based on the injective ReLU proposed by \citet{kothari2021}. We believe it to be of general interest as a dimension-changing conformal nonlinearity, but it provided no performance improvements in experiments. 

More useful was the \emph{conditional orthogonal} transformation, which takes advantage of the norm-preservation of orthogonal transformations to create an invertible layer. Despite being discontinuous, it provided a substantial boost in reconstruction ability on image data. The idea behind this transformation can be extended to the other parameterized mappings in Table \ref{tab:trans_rot_scale_sct}.
For each type of mapping we can identify hypersurfaces in $\mathbb{R}^n$ such that each hypersurface is mapped back to itself; i.e., each hypersurface is an orbit of its points under the mapping.
Applying the same type of conformal mapping piecewise on either side remains an invertible operation as long as trajectories do not cross the hypersurface, and the result is conformal \emph{almost everywhere}. The conditional orthogonal layer was the only example of these that provided performance improvements.

\section{Related Work}\label{sec:relatedwork}

\paragraph{Flows on prescribed manifolds.}
Flows can be developed for Riemannian manifolds $\cM \subseteq \cX$ which are known in advance and can be defined as the image of some fixed $\phi: \cU \to \cX$, where $\cU\subseteq \bR^m$ \cite{gemici2016, mathieu2020, papamakarios2021}. In particular, \citet{rezende2020} model densities on spheres and tori with convex combinations of M\"obius transformations, which are cognate to conformal mappings. For known manifolds $\phi$ is fixed, and the density's Jacobian determinant factor may be computable in closed form. Our work replaces $\phi$ with a trainable network $g$, but the log-determinant still has a simple closed form.

\paragraph{Flows on learnable manifolds.}
Extending flows to learnable manifolds brings about two main challenges: handling off-manifold points, and training the density on the manifold. 

When the distribution is manifold-supported, it will assign zero density to off-manifold points. This has been addressed by adding an off-manifold noise term \cite{cunningham2021, cunningham2020} or by projecting the data onto the manifold and training it with a reconstruction loss \cite{brehmer2020, kothari2021, kumar2020}. We opt for the latter approach.

Training the density on the manifold is challenging because the log-determinant term is typically intractable. \citet{kumar2020} use a series of lower bounds to train the log-determinant, while \citet{brehmer2020} and \citet{kothari2021} separate the flow into two components and train only the low-dimensional component. \citet{caterini2021} maximize log-likelihood directly by either constructing the embedding's Jacobian explicitly or using a stochastic approximation of the log-determinant's gradient, but both approaches remain computationally expensive for high-dimensional data. Our approach is the first injective model to provide a learnable manifold with exact and efficient log-determinant computation.

\paragraph{Conformal networks.}
Numerous past works have imposed approximate conformality or its special cases as a regularizer \cite{bansal2018, jia2016, peterfreund2020, qi2020, xiao2018}, but it has been less common to enforce conformality strictly. To maintain orthogonal weights, one must optimize along the \textit{Stiefel manifold} of orthogonal matrices. Past work to achieve this has either trained with Riemannian gradient descent or directly parameterized subsets of orthogonal matrices. Riemannian gradient descent algorithms typically require a singular value or QR decomposition at each training step \cite{harandi2016,huang2018, ozay2016}. We found that orthogonal matrices trained more quickly when directly parameterized. In particular, \citet{lezcano2019a} and \citet{lezcano2019b} parameterize orthogonal matrices as the matrix exponential of a skew-symmetric matrix, and \citet{tomczak2016} use Householder matrices. We used a mix of both.
\section{Experiments}\label{sec:experiments}

To implement CEFs, we worked off of the nflows github repo \cite{nflows}, which is derived from the code of \citet{durkan2019neural}. Our code is available at \href{https://github.com/layer6ai-labs/CEF}{\tt{https://github.com/layer6ai-labs/CEF}}. Full model and training details are provided in App. \ref{app:experiments}, while additional reconstructions and generated images are presented in App. \ref{app:ims}.

\subsection{Spherical Data}\label{sec:sphere}

\begin{figure}[h]
\centering
\begin{tabular}{cc}
    \includegraphics[width=\textwidth]{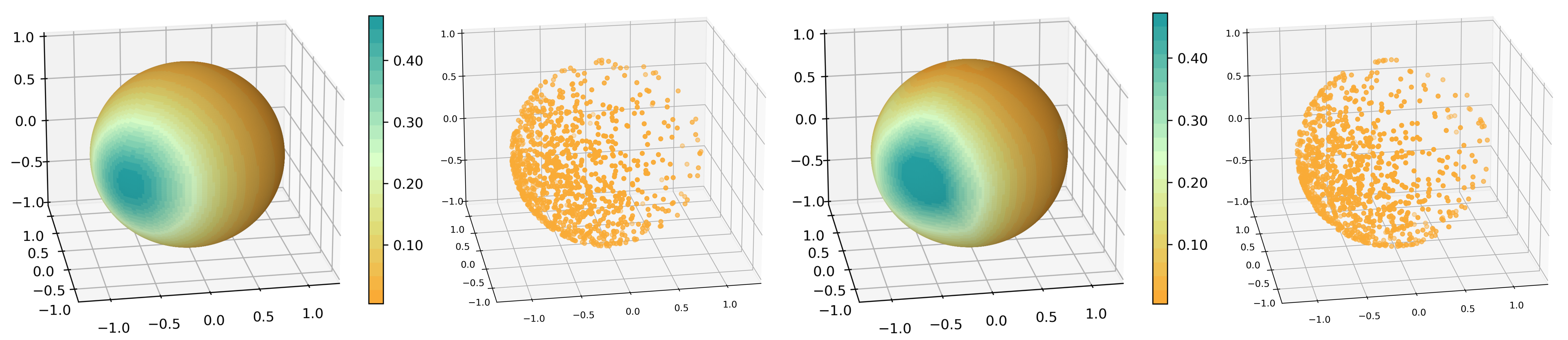} \\
    (a)\kern 95pt (b) \kern 85pt  (c) \kern 85pt (d)
\end{tabular}
\caption{(a) A density $p^*_{\mathbf{x}}(\mathbf{x})$ with support on the sphere, and (b) $10^3$ samples comprising the training dataset $\{\mathbf{x_i}\}$. (c) The density learned by a CEF, and (d) $10^3$ generated samples.}
\label{fig:sphere}
\end{figure}

To demonstrate how a CEF can jointly learn a manifold and density, we generated a synthetic dataset from a known distribution with support on a spherical surface embedded in $\mathbb{R}^3$ as described in App. \ref{app:experiments}. The distribution is visualized in Fig. \ref{fig:sphere}, along with the training dataset of $10^3$ sampled points.

We trained the two components of the CEF jointly, using the mixed loss function in Eq. \eqref{eq:mixed-loss-function} with an end-to-end log-likelihood term. The resulting model density is plotted in Fig. \ref{fig:sphere} along with generated samples. It shows good fidelity to the true manifold and density.

\subsection{Image Data}

We now evaluate manifold flows on image data. Our aim is to show that, although they represent a strict architectural subset of mainstream injective flows, CEFs remain competitive in generative performance \cite{brehmer2020, kothari2021}. In doing so, this work is the first to include end-to-end maximum likelihood training with an injective flow on image data. Three approaches were evaluated on each dataset: a jointly trained CEF, a sequentially trained CEF, and for a baseline a sequentially trained injective flow, as in \citet{brehmer2020}, labelled \textit{manifold flow} (MF).

Injective models cannot be compared on the basis of log-likelihood, since each model may have a different manifold support. Instead, we evaluate generative performance in terms of fidelity and diversity \cite{sajjadi2018}. The \textit{FID score} \cite{heusel2017} is a single metric which combines these factors, whereas \textit{density} and \textit{coverage} \cite{naeem2020} measure them separately. For FID, lower is better, while for density and coverage, higher is better. We use the PyTorch-FID package \cite{seitzer2020} and the implementation of density and coverage from \citet{naeem2020}.

\begin{figure}[]
\centering
    \includegraphics[width=0.7\textwidth]{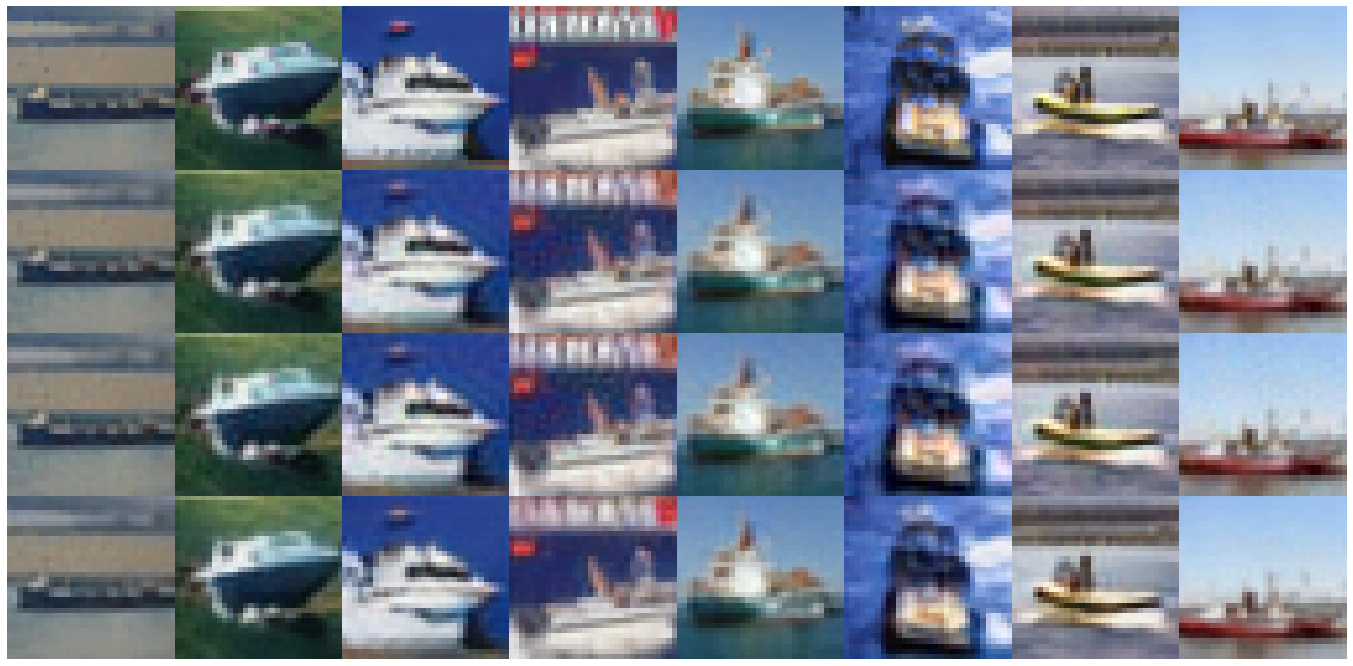}
\caption{Synthetic 512-dimensional Ship Manifold Reconstructions. From top to bottom: groundtruth samples, joint CEF, sequential CEF, and sequential MF.}
\label{fig:cifar10-512-recon}
\end{figure}

\begin{figure}
\centering
    \includegraphics[width=0.7\textwidth]{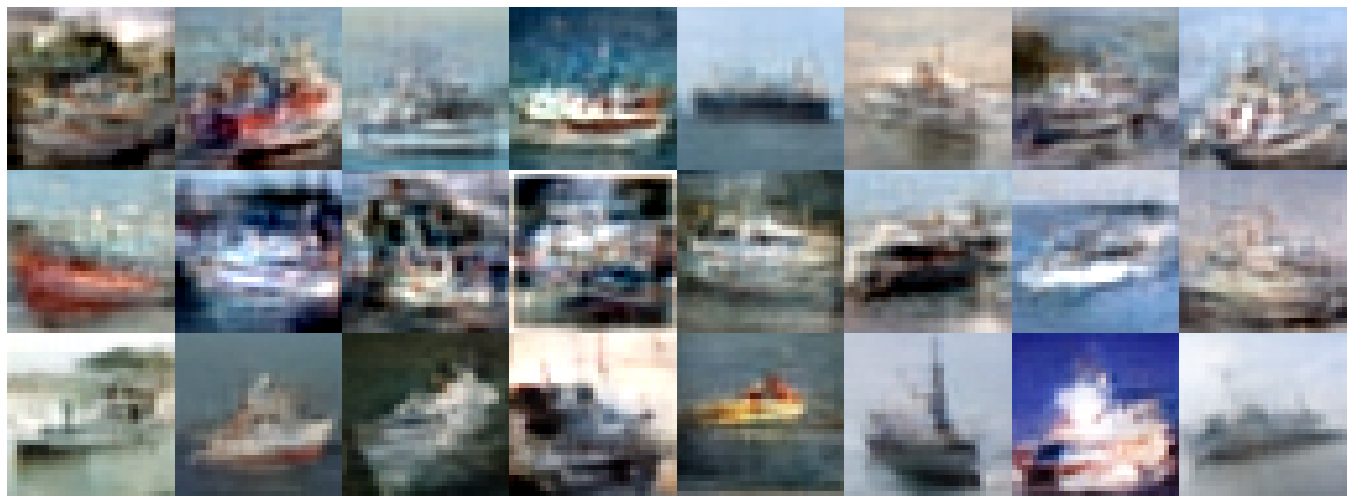}
\caption{Uncurated Synthetic 512-dimensional Ship Manifold Samples. From top to bottom: joint CEF, sequential CEF, and sequential MF.}
\label{fig:cifar10-samples}
\end{figure}

\paragraph{Synthetic image manifolds.}
Before graduating to natural high-dimensional data, we test CEFs on a high-dimensional synthetic data manifold whose properties are better understood. We generate data using a GAN pretrained on CIFAR-10 \cite{krizhevsky2009learning} by sampling from a selected number of latent space dimensions (64 and 512) with others held fixed. Specifically, we sample a single class from the class-conditional StyleGAN2-ADA provided by \citet{karras2020}. This setup reflects our model design in that (1) the true latent dimension is known and (2) since a single class is used, the resulting manifold is more likely to be connected. On the other hand, the GAN may not be completely injective, so its support may not technically be a manifold. Results are shown in Table \ref{tbl:synthetic_images}, Figs. \ref{fig:cifar10-512-recon} and \ref{fig:cifar10-samples}, and App. \ref{app:ims}.

\begin{table}[]
\small
\caption{Synthetic CIFAR-10 Ship Manifolds}
    \centering
    \begin{tabular}{l llll c llll}
        \toprule
        \multirow{2}{*}{\textsc{Model}} & \multicolumn{4}{c}{\textsc{64 dimensions}} &~& \multicolumn{4}{c}{\textsc{512 dimensions}} \\
        \cmidrule{2-5} \cmidrule{7-10}
         & \textsc{Recon} & \textsc{FID} & \textsc{Density} & \textsc{Cov} && \textsc{Recon} & \textsc{FID} & \textsc{Density} & \textsc{Cov}\\
        \midrule
        \textsc{J-CEF}  & 0.000695  & 36.5 & 0.0491 & 0.0658 && 0.000568 & 76.2 & 0.398 & 0.266\\
        \textsc{S-CEF}  &  0.000717 & 35.3 & 0.0548 & 0.0640 && 0.000627 & 74.7 & 0.421 & 0.251 \\
        \textsc{S-MF}  &  0.000469 & 28.7 & 0.0756 & 0.1103 && 0.000568 & 53.6 & 0.570& 0.446  \\
        \bottomrule
    \end{tabular}
    \vskip-0.3cm
\label{tbl:synthetic_images}
\end{table}

All models achieve comparable reconstruction losses with very minor visible artifacts, showing that conformal embeddings can learn complex manifolds with similar efficacy to state-of-the-art flows, despite their restricted architecture. The manifold flow achieves better generative performance based on our metrics and visual clarity. Between the CEFs, joint training allows the learned manifold to adapt better, but this did not translate directly to better generative performance.

\paragraph{Natural image data.} 

\begin{figure}
\centering
    \includegraphics[width=0.7\textwidth]{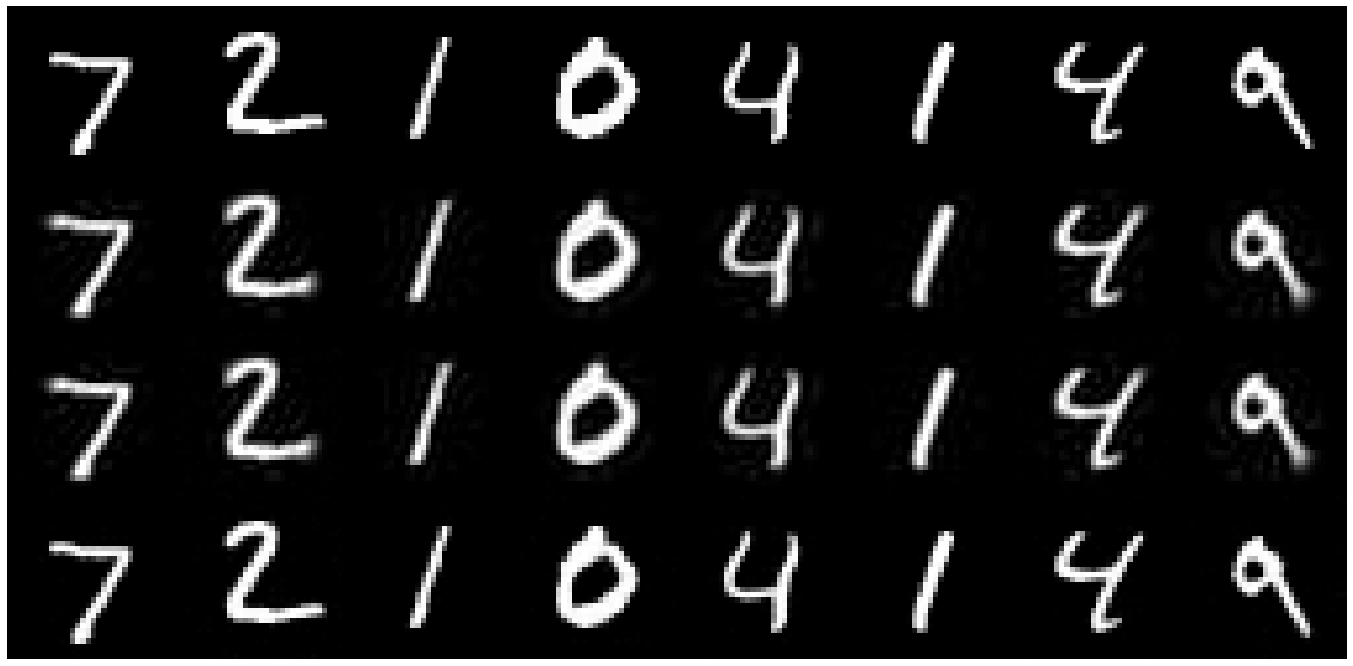}
\caption{MNIST Reconstructions. From top to bottom: groundtruth samples, joint CEF, sequential CEF, and sequential MF.}
\label{fig:mnist-recon}
\end{figure}

\begin{figure}
\centering
    \includegraphics[width=1.0\textwidth]{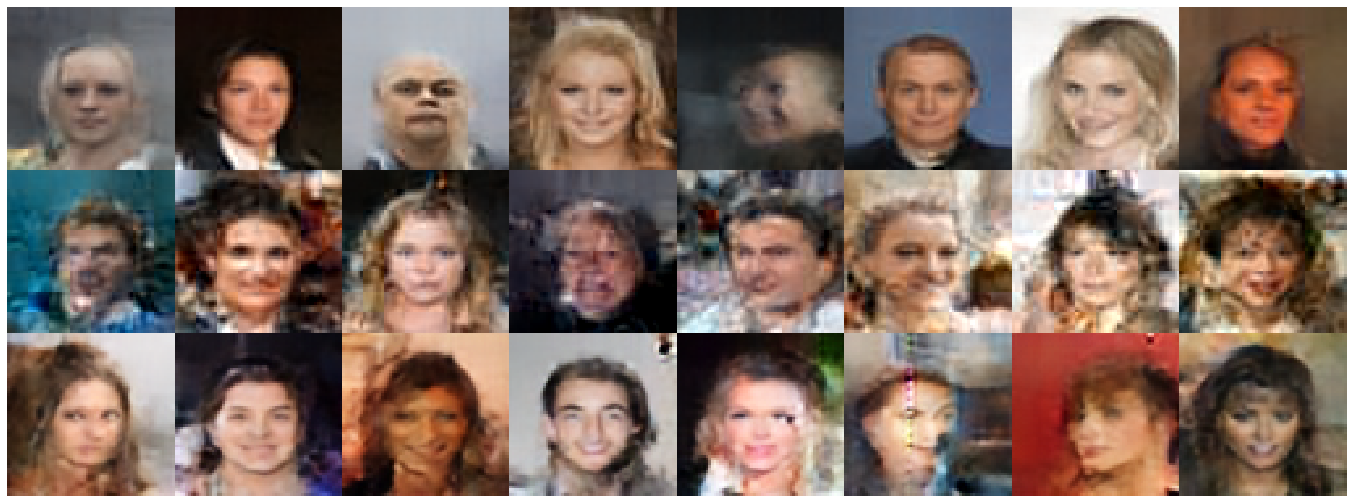}
\caption{Uncurated CelebA Samples. From top to bottom: joint CEF, sequential CEF, and sequential MF.}
\label{fig:celeba-samples}
\end{figure}

We scale CEFs to natural data by training on the MNIST \cite{lecun1998gradient} and CelebA \cite{liu2015} datasets, for which a low-dimensional manifold structure is postulated but unknown. Results are given in Table \ref{tbl:natural_images}, Figs. \ref{fig:mnist-recon} and \ref{fig:celeba-samples}, and App. \ref{app:ims}.

As expected, since the MF's embedding is more flexible, it achieves smaller reconstruction losses than the CEFs. On MNIST, this is visible as faint blurriness in the CEF reconstructions in Fig. \ref{fig:mnist-recon}, and it translates to better sample quality for the MF as per the metrics in Table \ref{tbl:natural_images}. Interestingly however, the jointly-trained CEF obtains substantially better sample quality on CelebA, both visually (Fig. \ref{fig:celeba-samples}) and by every metric. We posit this is due to the phenomenon observed by \citet{caterini2021} in concurrent work: for complex distributions, the learned manifold parameterization has significant influence on the difficulty of the density estimation task. Only the joint training approach, which maximizes likelihoods end-to-end, can train the manifold parameterization to an optimal starting point for density estimation, while sequential training optimizes the manifold solely on the basis of reconstruction loss. CelebA is the highest-dimensional dataset tested here, and its distribution is presumably quite complex, so one can reasonably expect joint training to provide better results. On the other hand, the sequentially trained CEF's performance suffers from the lack of both joint training and the expressivity afforded by the more general MF architecture.

\begin{table}[]
\small
\caption{Natural Image Data}
    \centering
    \begin{tabular}{l llll  c llll}
        \toprule
        \multirow{2}{*}{\textsc{Model}} & \multicolumn{4}{c}{\textsc{MNIST}} &~& \multicolumn{4}{c}{\textsc{CelebA}} \\
        \cmidrule{2-5} \cmidrule{7-10}
         & \textsc{Recon} & \textsc{FID} & \textsc{Density} & \textsc{Cov} && \textsc{Recon} & \textsc{FID} & \textsc{Density} & \textsc{Cov}\\
        \midrule
        \textsc{J-CEF}  & 0.003222 & 38.5 & 0.0725 & 0.1796 && 0.001016  &   118 &  0.05581 & 0.00872\\
        \textsc{S-CEF}  &  0.003315 & 37.9 & 0.0763& 0.1800 &&  0.001019 &   171 &  0.00922 & 0.00356\\
        \textsc{S-MF}  & 0.000491 & 16.1 & 0.5003 & 0.7126 &&  0.000547 &   142 &  0.02425 & 0.00576\\
        \bottomrule
    \end{tabular}
    \vskip-0.3cm
\label{tbl:natural_images}
\end{table}

\section{Limitations and Future Directions}\label{sec:limitations}

\paragraph{Expressivity.}
Just as standard flows trade expressivity for tractable likelihoods, so must injective flows. Our conformal embeddings in particular are less expressive than state-of-the-art flow models; they had higher reconstruction loss than the neural spline flow-based embeddings we tested. The conformal embeddings we designed were limited in that they mostly derive from dimension-\textit{preserving} conformal mappings, which is a naturally restrictive class by Liouville's theorem \cite{hartman1958}. Just as early work on NFs \cite{dinh2014,rezende2015} introduced limited classes of parameterizable bijections, which were later improved substantially (e.g. \cite{durkan2019neural,kingma2018}), our work introduces several classes of parameterizable conformal embeddings. We expect that future work will uncover more expressive conformal embeddings.

\paragraph{Manifold learning.}
Strictly manifold-supported probability models such as ours introduce a bi-objective optimization problem. How to balance these objectives is unclear and, thus far, empirical \cite{brehmer2020}. The difference in supports between two manifold models also makes their likelihoods incomparable. \citet{cunningham2020} have made progress in this direction by convolving the manifold-supported distribution with noise, but this makes inference stochastic and introduces density estimation challenges. We suspect that using conformal manifold-learners may make density estimation more tractable in this setting, but further research is needed in this direction.

\paragraph{Broader impact.}
As deep generative models become more advanced, researchers should carefully consider some accompanying ethical concerns. Large-scale, natural image datasets carry social biases which are likely to be codified in turn by the models trained on them \cite{steed2021image}. For instance, CelebA does not accurately represent the real-world distribution of human traits, and models trained on CelebA should be vetted for fairness before being deployed to make decisions that can adversely affect people. Deep generative modelling also lends itself to malicious practices \cite{brundage2021malicious} such as disinformation and impersonation using deepfakes \cite{westerlund2019emergence}.

Our work seeks to endow normalizing flows with more realistic assumptions about the data they model. While such improvements may invite malicious downstream applications, they also encode a better understanding of the data, which makes the model more interpretable and thus more transparent. We hope that a better understanding of deep generative models will synergize with current lines of research aimed at applying them for fair and explainable real-world use \cite{balunovic2021fair,nalisnick2018deep}.
\section{Conclusion}\label{sec:conclusion}

This paper introduced Conformal Embedding Flows for modelling probability distributions on low-dimensional manifolds while maintaining tractable densities. We showed that conformal embeddings naturally match the framework of normalizing flows by providing efficient sampling, inference, and density estimation, and they are composable so that they can be scaled to depth. Furthermore, it appears conformality is a minimal restriction in that any looser condition will sacrifice one or more of these properties. As we have reviewed, previous instantiations of injective flows have not maintained all of these properties simultaneously.

Normalizing flows are still outperformed by other generative models such as GANs and VAEs in the arena of realistic image generation. Notably, these two alternatives benefit from a low-dimensional latent space, which better reflects image data's manifold structure and provides for more scalable model design. By equipping flows with a low-dimensional latent space, injective flow research has made progress towards VAE- or GAN-level performance. The CEF paradigm is a way to match these strides while maintaining the theoretical strengths of NFs.

\begin{ack}
We thank Gabriel Loaiza-Ganem and Anthony Caterini for their valuable discussions and advice. We also thank Parsa Torabian for sharing his experience with orthogonal weights and Maksims Volkovs for his helpful feedback.

The authors declare no competing interests or third-party funding sources.
\end{ack}

\bibliographystyle{abbrvnat}
\bibliography{main.bib}{}

\newpage

\appendix

\section{Injective Flows are Manifold Learners}\label{app:recon}

\subsection{Densities on Manifolds}\label{app:manifold-learning}

In this work we model a probability density $p$ on a Riemannian manifold $\mathcal{M}$. Here, we briefly review what this means formally.

Consider a probability measure $\mathbb{P}$ on the space $\mathcal{X} = \mathbb{R}^n$.
We say $\mathbb{P}$ admits a density with respect to a base measure $\mu$ if $\mathbb{P}$ is absolutely continuous with respect to $\mu$. If so, we let the density be $p := \frac{d\mathbb{P}}{d\mu}$ (the Radon-Nikodym derivative of $\mathbb{P}$ with respect to $\mu$). The base measure $\mu$ is most commonly the Lebesgue measure on $\mathcal{X}$.

However, if $\mathbb{P}$ is supported on an $m$-dimensional Riemannian submanifold $\mathcal{M} \subset \mathcal{X}$, some adjustment is required. $\mathbb{P}$ will not be absolutely continuous with respect to the Lebesgue measure on $\mathcal{X}$, so we require a different choice of base measure.
In the literature involving densities on manifolds (see Sec. \ref{sec:relatedwork}), it is always assumed, but seldom explicitly stated, that the density's base measure is the Riemannian measure $\mu$ \cite{pennec2006intrinsic} of the submanifold $\mathcal{M}$. Furthermore, the Riemmanian metric from which $\mu$ arises is always inherited from the Euclidean metric of ambient space $\mathcal{X}$. From this construction we gather that, whenever a Riemannian submanifold $\mathcal{M} \subset \mathcal{X}$ is specified, a unique natural base measure $\mu$ follows.

Given an injective flow model $\mathbf{f}_\theta : \cZ \to \cX$ with latent space $\cZ = \bR^m$, the goal is to make the implied model manifold $\mathcal{M}_\theta = \mathbf{f}_\theta(\cZ)$ match the data manifold $\mathcal{M}_d$. As per the previous paragraph, this task equates to learning the base measure on which the model's density will be evaluated, which is necessarily a separate objective from likelihood maximization. Below we discuss how minimizing reconstruction loss achieves the goal of manifold matching.

\subsection{Reconstruction Minimization for Manifold Learning}\label{app:recon-proof}

Let $\mathbf{f}_\theta: \cZ \to \cX=\mathbb{R}^n$ be an injective flow ($m < n$) with a smooth left-inverse $\mathbf{f}_\theta^\dagger$ as described in Sec. \ref{sec:background_injective}. By construction, the image $\mathcal{M}_\theta = \mathbf{f}_\theta(\cZ)$ of the flow is a Riemannian submanifold of $\cX$; we call $\mathcal{M}_\theta$ the \textit{model manifold}. Let $P$ be a data distribution supported by an $m$-dimensional Riemannian submanifold $\mathcal{M}_d$; we call this the \textit{data manifold}. Suppose furthermore that $P$ admits a probability density with respect to the Riemannian measure of $\mathcal{M}_d$.

\paragraph{Proposition 1} \emph{$\cR(\theta) := \mathbb{E}_{\bx \sim P}||\bx - \mathbf{f}_\theta(\mathbf{f}_\theta^\dagger(\bx))||^2 = 0$ if and only if $\mathcal{M}_d \subseteq \text{cl}\left(\mathcal{M}_\theta\right)$.} 

Since the reconstruction loss $\cR(\theta)$ is continuous, we infer that $\cR(\theta) \to 0$ will bring $\mathcal{M}_\theta$ and $\mathcal{M}_d$ into alignment (except possibly for the $P$-null set $\mathcal{M}_d \cap \partial\mathcal{M}_\theta$).

\paragraph{Proof}
For the forward direction, suppose $\cR(\theta) = 0$. For a single point $\bx \in \mathbb{R}^n$, by the definition of $\mathbf{f}_\theta^\dagger$ we have $\bx = \mathbf{f}_\theta( \mathbf{f}_\theta^\dagger(\bx))$ if and only if $\bx \in \mathbf{f}_\theta(\cZ) = \mathcal{M}_\theta$. Put differently,  $\mathcal{M}_\theta = \{ \bx \in \cX: \bx = \mathbf{f}_\theta( \mathbf{f}_\theta^\dagger(\bx) ) \}$, so a reconstruction error of zero implies $P(\mathcal{M}_\theta) = 1$. This means that $P$'s support must by definition be a subset of $\text{cl}\left(\mathcal{M}_\theta\right)$. It follows that $\mathcal{M}_d \subseteq \text{cl}\left(\mathcal{M}_\theta\right)$.

For the reverse direction, note that if $\mathcal{M}_d \subseteq \text{cl}\left(\mathcal{M}_\theta\right)$, then $\mathcal{M}_d \setminus \mathbf{f}_\theta(\cZ)$ has measure zero in $\mathcal{M}_d$, so $\mathbf{f}_\theta(\mathbf{f}_\theta^\dagger(\bx)) = 0$ $P$ -almost surely. This fact yields $\cR(\theta) = 0$.
\qed

\subsection{Joint Training and Wasserstein Training}\label{app:wasserstein}

The Wasserstein-1 distance between the groundtruth and model distributions is another objective motivated by the low-dimensional manifold structure of high-dimensional data. In some sense, minimizing Wasserstein-1 distance is a more elegant approach than sequential or joint training because it is a distance metric between probability distributions, whereas the sequential and joint objectives involve separate terms for the support and the likelihood within. However, in practice the Wasserstein distance cannot be estimated without bias in a polynomial number of samples \cite{arora2017generalization}, and it must be estimated adversarially \cite{arjovsky2017wasserstein,tolstikhin2018wasserstein}.

We compare our joint training method to Wasserstein training on 64000 points sampled from a 2D Gaussian mixture embedded as a plane in 3D space. The conformal embedding $\bg$ is a simple orthogonal transformation from 2 into 3 dimensions, which makes the manifold easy to plot. We let $\bh$ be a simple flow consisting of 3 LU-decomposed linear transformations interspersed with coupling layers, where each layer has 2 residual blocks with 128 hidden features. Both affine coupling and rational-quadratic coupling layers were tested.

\begin{figure}[]
\centering
\includegraphics[trim=60 20 15 80, clip,width=0.4\textwidth]{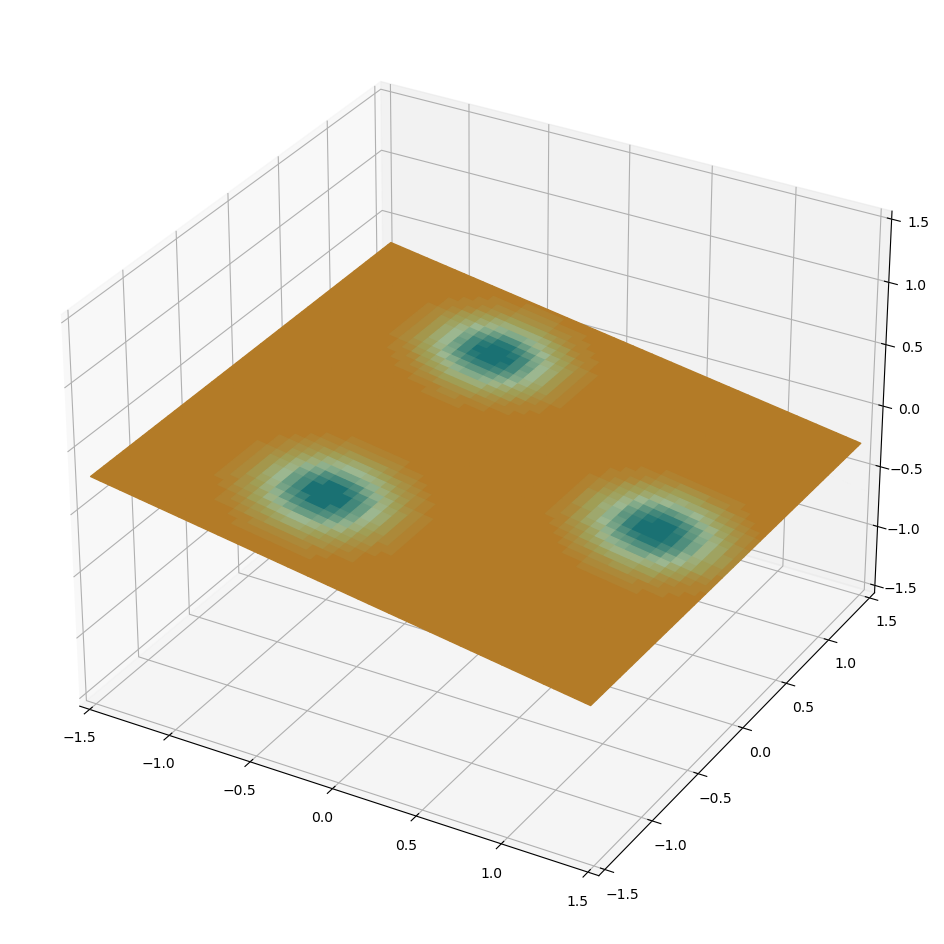}
\begin{tabular}{cc}
    \includegraphics[trim=50 20 15 80, clip,width=0.4\textwidth]{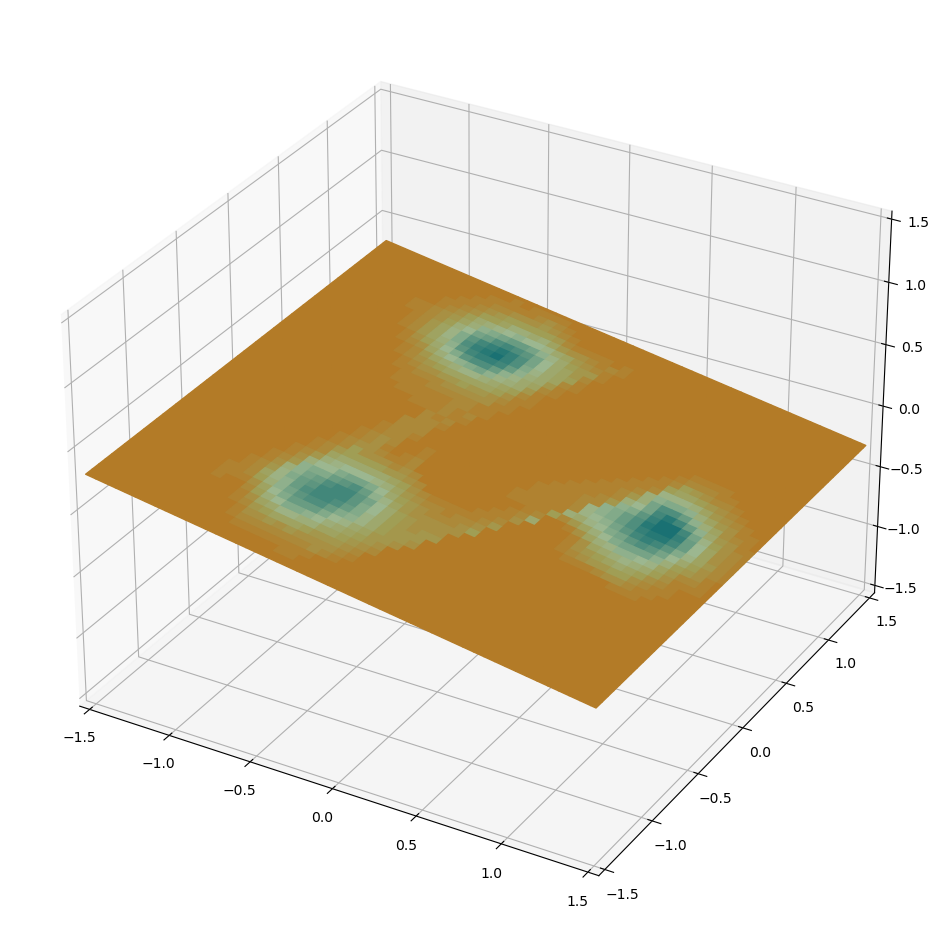}
    \includegraphics[trim=40 20 10 55, clip,width=0.4\textwidth]{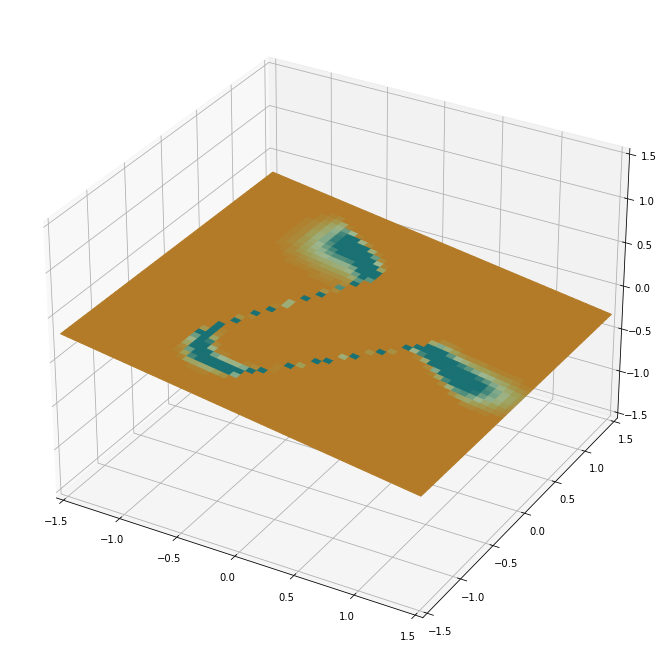}\\
    \includegraphics[trim=50 20 15 80, clip,width=0.4\textwidth]{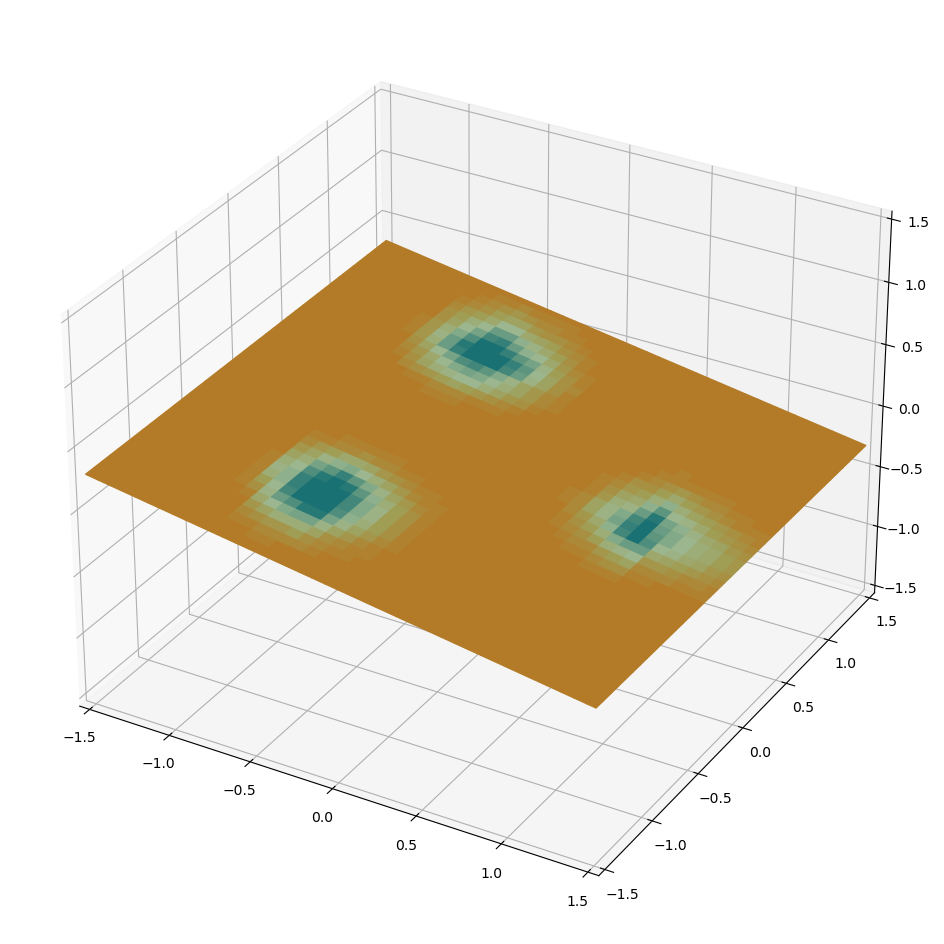}
    \includegraphics[trim=50 20 15 80, clip,width=0.4\textwidth]{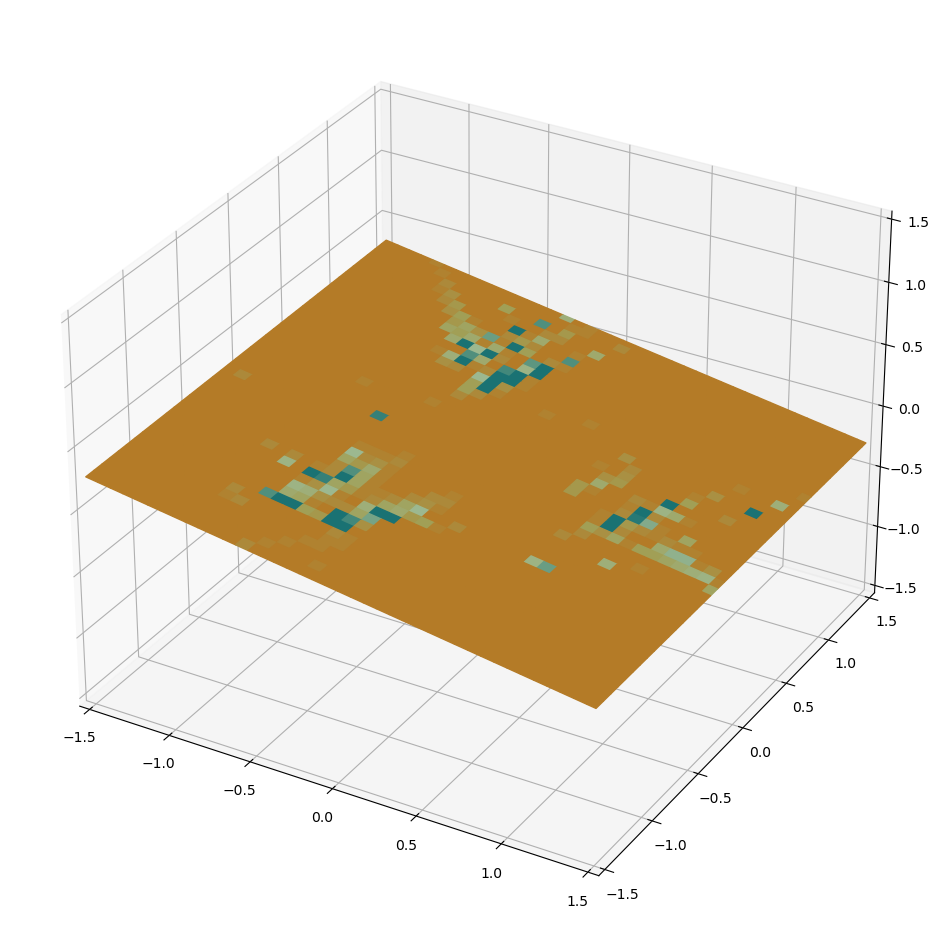}
\end{tabular}
\caption{Gaussian mixture on a 2D plane. From top to bottom: groundtruth distribution, the learned distribution using affine coupling layers, and learned distribution using rational-quadratic coupling layers. The results of joint training are shown on the left, while the results of Wasserstein adversarial training are shown on the right.}
\label{fig:mixture}
\end{figure}

Each architecture was separately trained with both the joint loss and with adversarially estimated Wasserstein loss. The discriminator was a 16-hidden layer ReLU MLP with 512 hidden units each. We enforce the Lipschitz constraint using gradient penalties \cite{gulrajani2017improved}. Results are visible in Fig. \ref{fig:mixture}.

When adversarially trained, the inductive bias of the generator appears to play a strong role, and neither model learns the density well. These results suggest that training a flow jointly provides better density estimation than estimating the Wasserstein loss. Our observations support those of \citet{grover2018flowgan}, who show that adversarial training can be counterproductive to likelihood maximization, and \citet{brehmer2020}, who report poor results when their model is trained for optimal transport.

\section{Details on Conformal Embeddings and Conformal Mappings}\label{app:conformal}

Let $(\cU, \eta_\bu)$ and $(\cX, \eta_\bx)$ be two Riemannian manifolds. We define a diffeomorphism $\mathbf{f}: \cU \to \cX$ to be a \textit{conformal diffeomorphism} if it pulls back the metric $\eta_\bx$ to be some non-zero scalar multiple of $\eta_\bu$ \cite{lee2018}. That is,
\begin{equation}
    \mathbf{f}^* \eta_\bx = \lambda^2 \eta_\bu
\end{equation}
for some smooth non-zero scalar function $\lambda$ on $\cU$. Furthermore, we define a smooth embedding $\bg: \cU \to \cX$ to be a \textit{conformal embedding} if it is a conformal diffeomorphism onto its image $(\bg(\cU), \eta_\bx)$, where $\eta_\bx$ is inherited from the ambient space $\cX$.

In our context, $\cU \subseteq \bR^m$, $\cX = \bR^n$, and $\eta_\bu$ and $\eta_\bx$ are Euclidean metrics. This leads to an equivalent property (Eq. \eqref{eq:conformal_transformation}):
\begin{equation}\label{eq:conformal_transformation_repeat}
    \mathbf{J}_{\mathbf{g}}^T(\mathbf{u}) \mathbf{J}_{\mathbf{g}}(\mathbf{u}) = \lambda^2(\mathbf{u})\mathbf{I}_m.
\end{equation}
This also guarantees that $\det [\mathbf{J}_{\mathbf{g}}^T \mathbf{J}_{\mathbf{g}}] = \lambda^{2m}$ is tractable, even when $\bg=\bg_k\circ ...\circ\bg_1$ is composed from several layers, as is needed for scalable injective flows.

To demonstrate that conformal embeddings are an expressive class of functions, we first turn to the most restricted case where $n=m$; i.e. conformal mappings. In Apps. \ref{app:conformal_transformations} and \ref{app:finite-conformal} we provide an intuitive investigation of the classes of conformal mappings using infinitesimals. We then discuss in App. \ref{app:conformal_embeddings} why conformal embeddings in general are more challenging to analyze, but also show intuitively why they are more expressive than dimension-preserving conformal mappings.

\subsection{Infinitesimal Conformal Mappings}\label{app:conformal_transformations}

Consider a mapping of Euclidean space with dimension $m \geq 3$. Liouville's theorem for conformal mappings constrains the set of such maps which satisfy the conformal condition Eq. \eqref{eq:conformal_transformation_repeat}. Such functions can be decomposed into translations, orthogonal transformations, scalings, and inversions. Here we provide a direct approach for the interested reader, which also leads to some insight on the general case of conformal embeddings \cite{francesco2012}. First we will find all infinitesimal transformations which satisfy the conformal condition, then exponentiate them to obtain the set of finite conformal mappings.

Consider a transformation $\mathbf{f}: \mathbb{R}^m\to \mathbb{R}^m$ which is infinitesimally close to the identity function, expressed in Cartesian coordinates as
\begin{equation}\label{eq:infinitesimal_function}
    \mathbf{f}(\mathbf{x}) =  \mathbf{x} + \boldsymbol{\epsilon}(\mathbf{x}).
\end{equation}
That is, we only keep terms linear in the infinitesimal quantity $\boldsymbol{\epsilon}$. The mappings produced will only encompass transformations which are continuously connected to the identity, but we restrict our attention to these for now. However, this simple form allows us to directly study how Eq. \eqref{eq:conformal_transformation_repeat} constrains the infinitesimal $\boldsymbol{\epsilon}(\mathbf{x})$:
\begin{align}
\begin{aligned}
      \mathbf{J}^T_{\mathbf{f}}(\mathbf{x}) \mathbf{J}_{\mathbf{f}}(\mathbf{x}) &= \left[\mathbf{I}_m + \frac{\partial\boldsymbol{\epsilon}}{\partial \mathbf{x}}\right]^T\left[\mathbf{I}_m + \frac{\partial\boldsymbol{\epsilon}}{\partial \mathbf{x}}\right] \\
     &= \mathbf{I}_m + \frac{\partial\boldsymbol{\epsilon}}{\partial \mathbf{x}}^T + \frac{\partial\boldsymbol{\epsilon}}{\partial \mathbf{x}}.
     \end{aligned}
\end{align}
By Eq. \eqref{eq:conformal_transformation_repeat}, the symmetric sum of $\partial\boldsymbol{\epsilon} / {\partial \mathbf{x}}$ must be proportional to the identity matrix. Let us call the position-dependent proportionality factor $\eta(\mathbf{x})$. We can start to understand $\eta(\mathbf{x})$ by taking a trace
\begin{align}
     \frac{\partial\boldsymbol{\epsilon}}{\partial \mathbf{x}}^T + \frac{\partial\boldsymbol{\epsilon}}{\partial \mathbf{x}} & = \eta(\mathbf{x})\mathbf{I}_m,\label{eq:derivation1} \\
     \frac{2}{m}\text{tr}\left(\frac{\partial\boldsymbol{\epsilon}}{\partial \mathbf{x}}\right)&=  \eta(\mathbf{x}).\label{eq:derivation6}
\end{align}
Taking another derivative of Eq. \eqref{eq:derivation1} proves to be useful, so we switch to index notation to handle the tensor multiplications,
\begin{align}
     \frac{\partial}{\partial x_k}\frac{\partial{\epsilon_j}}{\partial {x_i}} +  \frac{\partial}{\partial x_k}\frac{\partial{\epsilon_i}}{\partial {x_j}} & = \frac{\partial{\eta}}{\partial {x_k}}\delta_{ij},
\end{align}
where the Kronecker delta $\delta_{ij}$ is 1 if $i=j$, and 0 otherwise.
On the left-hand-side, derivatives can be commuted. By taking a linear combination of the three permutations of indices we come to 
\begin{align}\label{eq:derivation5}
     2 \frac{\partial}{\partial x_k}\frac{\partial{\epsilon_i}}{\partial {x_j}} & = \frac{\partial{\eta}}{\partial {x_j}}\delta_{ik} + \frac{\partial{\eta}}{\partial {x_k}}\delta_{ij} - \frac{\partial{\eta}}{\partial {x_i}}\delta_{jk}.
\end{align}
Summing over elements where $j=k$ gives the Laplacian of $\epsilon _i$, while picking up only the derivatives of $\eta$ with respect to $x_i$, so we can switch back to vector notation where
\begin{align}\label{eq:derivation2}
     2 \nabla^2 \boldsymbol\epsilon & = (2 - m)\frac{\partial{\eta}}{\partial {\mathbf{x}}}.
\end{align}
Now we have two equations \eqref{eq:derivation1} and \eqref{eq:derivation2}\footnote{We note that the steps following Eq. \eqref{eq:derivation2} are only justified for $m\geq 3$ which we have assumed. In two dimensions the conformal group is much larger and Liouville's theorem no longer captures all conformal mappings.} involving derivatives of $\boldsymbol\epsilon$ and $\eta$. To eliminate $\boldsymbol\epsilon$, we can apply $\nabla^2$ to  \eqref{eq:derivation1}, while applying $\partial/\partial\mathbf{x}$ to \eqref{eq:derivation2}
\begin{align}
\nabla^2\frac{\partial\boldsymbol{\epsilon}}{\partial \mathbf{x}}^T + \nabla^2\frac{\partial\boldsymbol{\epsilon}}{\partial \mathbf{x}} & = \nabla^2\eta\mathbf{I}_d\label{eq:derivation3} \\
     2 \nabla^2 \frac{\partial\boldsymbol{\epsilon}}{\partial \mathbf{x}} & = (2 - m)\frac{\partial^2{\eta}}{\partial {\mathbf{x}}\partial {\mathbf{x}}}.\label{eq:derivation4}
\end{align}
Since Eq. \eqref{eq:derivation4} is manifestly symmetric, the left-hand-sides are actually equal. Equating the right-hand-sides, we can again sum the diagonal terms, giving the much simpler form
\begin{align}
(m - 1)\nabla^2\eta = 0.
\end{align}
Ultimately, revisiting Eq. \eqref{eq:derivation4} shows that the function $\eta(\mathbf{x})$ is linear in the coordinates
\begin{equation}
    \frac{\partial^2{\eta}}{\partial {\mathbf{x}}\partial {\mathbf{x}}} = 0  \implies \eta(\mathbf{x}) = \alpha+\boldsymbol\beta\cdot\mathbf{x},
\end{equation}
for constants $\alpha, \boldsymbol\beta$. This allows us to relate back to the quantity of interest $\boldsymbol\epsilon$. Skimming back over the results so far, the most general equation where having the linear expression for $\eta(\mathbf{x})$ helps is Eq. \eqref{eq:derivation5} which now is
\begin{align}\label{eq:derivation7}
     2 \frac{\partial}{\partial x_k}\frac{\partial{\epsilon_i}}{\partial {x_j}} & = \beta_j\delta_{ik} + \beta_k\delta_{ij} -\beta_i\delta_{jk}.
\end{align}
The point is that the right-hand-side is constant, meaning that $\boldsymbol\epsilon(\mathbf{x})$ is at most quadratic in $\mathbf{x}$. Hence, we can make an ansatz for $\boldsymbol\epsilon$ in full generality, involving sets of infinitesimal constants
\begin{align}
     \boldsymbol\epsilon = \mathbf{a} + \mathbf{Bx} + \mathbf{x}\tensor{\mathbf{C}} \mathbf{x},
\end{align}
where $\tensor{\mathbf{C}}\in \mathbb{R}^{m\times m\times m}$ is a 3-tensor.

So far we have found that infinitesimal conformal transformations can have at most quadratic dependence on the coordinates. It remains to determine the constraints on each set of constants $\mathbf{a}$, $\mathbf{B}$, and $\tensor{\mathbf{C}}$, and interpret the corresponding mappings. We consider each of them in turn.

All constraints on $\boldsymbol\epsilon$ involve derivatives, so there is nothing more to say about the constant term. It represents an infinitesimal translation
\begin{equation}
     \mathbf{f}(\mathbf{x}) =  \mathbf{x} + \mathbf{a}.
\end{equation}
On the other hand, the linear term is constrained by Eqs. \eqref{eq:derivation1} and \eqref{eq:derivation6} which give
\begin{equation}\label{eq:derivation9}
    \mathbf{B} + \mathbf{B}^T = \frac{2}{m}\text{tr}(\mathbf{B})\mathbf{I}_m.
\end{equation}
Hence, $\mathbf{B}$ has an unconstrained anti-symmetric part $\mathbf{B}_\text{AS} = \tfrac{1}{2}(\mathbf{B} - \mathbf{B}^T)$ representing an infinitesimal rotation
\begin{equation}\label{eq:derivation10}
     \mathbf{f}(\mathbf{x}) =  \mathbf{x} + \mathbf{B}_\text{AS}\mathbf{x},
\end{equation}
while its symmetric part is diagonal as in Eq. \eqref{eq:derivation9},
\begin{equation}\label{eq:derivation9.5}
     \mathbf{f}(\mathbf{x}) =  \mathbf{x} + \lambda\mathbf{x},\quad \lambda =  \frac{1}{m}\text{tr}(\mathbf{B}),
\end{equation}
which is an infinitesimal scaling. This leaves only the quadratic term for interpretation which is more easily handled in index notation, i.e. $\epsilon_i = \sum_{lm}C_{ilm}x_l x_m$. The quadratic term is significantly restricted by Eq. \eqref{eq:derivation7},
\begin{align}\label{eq:derivation8}
\begin{aligned}
    2 \frac{\partial^2{}}{\partial x_k\partial {x_j}}\sum_{lm}C_{ilm}x_l x_m &= 2C_{ijk} = \beta_j\delta_{ik} + \beta_k\delta_{ij} -\beta_i\delta_{jk}.
    \end{aligned}
\end{align}
This allows us to isolate $\beta_k$ in terms of $C_{ijk}$, specifically from the trace over $C$'s first two indices,
\begin{align}
    2\sum_{i=j}C_{ijk} & = \beta_k + \beta_k m -\beta_k = \beta_k m.
\end{align}
Hereafter we use $b_k = \beta_k / 2 = \sum_{i=j}C_{ijk} / m$. Then with Eq. \eqref{eq:derivation8} the corresponding infinitesimal transformation is
\begin{align}\label{eq:derivation11}
    \begin{aligned}
     f_i(\mathbf{x}) &=  x_i + \sum_{jk}C_{ijk}x_j x_k \\
     &= x_i + \sum_{jk}(b_j \delta_{ik} + b_k\delta_{ij} - b_i \delta_{jk})x_j x_k \\
     &= x_i + 2 x_i\sum_j b_j x_j - b_i\sum_j (x_j)^2,\\
     \mathbf{f}(\mathbf{x}) &=  \mathbf{x} + 2 (\mathbf{b}\cdot\mathbf{x}) \mathbf{x} - \Vert \mathbf{x}\Vert ^2\mathbf{b}.
\end{aligned}
\end{align}

We postpone the interpretation momentarily.

Thus we have found all continuously parametrizable infinitesimal conformal mappings connected to the identity and showed they come in four distinct types. By composing infinitely many such transformations, or ``exponentiating" them, we obtain finite conformal mappings. Formally, this is the process of exponentiating the elements of a Lie algebra to obtain elements of a corresponding Lie group.

\subsection{Finite Conformal Mappings}\label{app:finite-conformal}

As an example of obtaining finite mappings from infinitesimal ones we take the infinitesimal rotations from Eq. \eqref{eq:derivation10} where we note that $\mathbf{f}$ only deviates from the identity by an infinitesimal vector field $\mathbf{B}_\text{AS}\mathbf{x}$. By integrating the field we get the finite displacement of any point under many applications of $\mathbf{f}$, i.e. the integral curves $\mathbf{x}(t)$ defined by
\begin{equation}
    \dot{\mathbf{x}}(t)=\mathbf{B}_{\text{AS}}\mathbf{x}(t), \quad \bx(0) = \bx_0.
\end{equation}
This differential equation has the simple solution
\begin{equation}
    \mathbf{x}(t)=\exp(t\mathbf{B}_{\text{AS}} )\mathbf{x}_0.
\end{equation}

\begin{figure}
\centering
\begingroup 
\setlength{\tabcolsep}{4pt}
\renewcommand{\arraystretch}{0.3} 
\begin{tabular}{c} 
    \begin{tabular}{cc}
    \includegraphics[width=34mm]{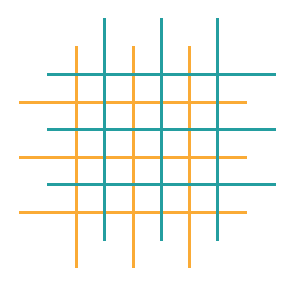} \hspace{-4mm} \includegraphics[width=34mm]{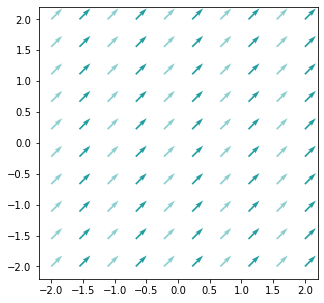} &
    \includegraphics[width=34mm]{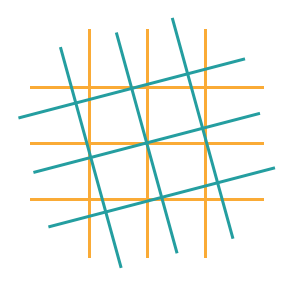} \hspace{-3mm} \includegraphics[width=34mm]{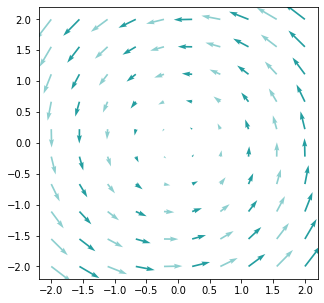}
    \\
    (a) & (b)
    \end{tabular}
    \\
    \begin{tabular}{cc}
    \includegraphics[width=34mm]{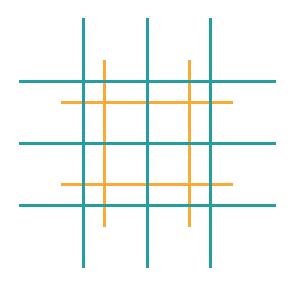} \hspace{-4mm} \includegraphics[width=34mm]{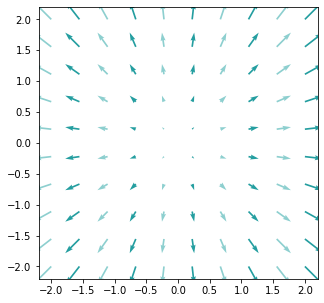} &
    \includegraphics[width=34mm]{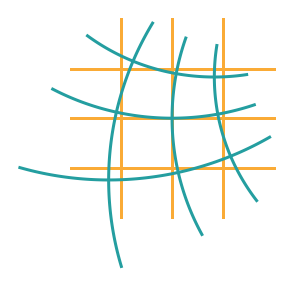} \hspace{-3mm} \includegraphics[width=34mm]{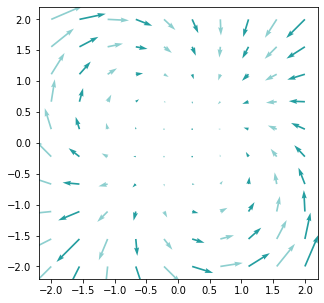}
    \\
    (c) & (d)
    \end{tabular}
    \\
    \begin{tabular}{c}
    \includegraphics[width=34mm]{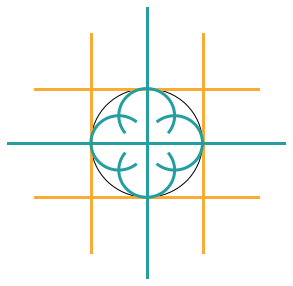} \hspace{-3mm} \includegraphics[width=34mm]{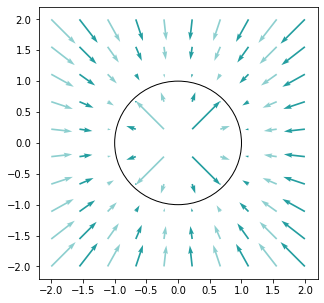}
    \\
    (e)
    \end{tabular}
\end{tabular}
\endgroup
\caption{Effects of conformal mappings on gridlines, and their  corresponding vector fields showing local displacements. Mappings are: (a) translation by $\mathbf{a}=[1,1]$; (b) orthogonal transformation (2D rotation) by angle $\theta=\pi/12$; (c) scaling by $\lambda = 1.5$; (d) SCT by $\mathbf{b}=[-0.1,-0.1]$; (e) inversion, also showing the unit circle. The interior of the circle is mapped to the exterior, and vice versa.}
\label{fig:conf_mappings}
\end{figure}

Finally we recognize that when a matrix $\mathbf{A}$ is antisymmetric, the matrix exponential $e^\mathbf{A}$ is orthogonal, showing that the finite transformation given by $t=1$, $\mathbf{f}(\mathbf{x}_0) = \exp(\mathbf{B}_{\text{AS}} )\mathbf{x}_0$, is indeed a rotation. Furthermore, it is intuitive that infinitesimal translations and scalings also compose into finite translations and scalings. Examples are shown in Fig. \ref{fig:conf_mappings} (a-c)

The infinitesimal transformation in Eq. \eqref{eq:derivation11} is non-linear in $\mathbf{x}$, so it does not exponentiate easily as for the other three cases. It helps to linearize with a change of coordinates $\mathbf{y} = \mathbf{x}/\Vert\mathbf{x}\Vert^2$ which happens to be an inversion:
\begin{align}
    \dot{\mathbf{x}}(t)&=2(\mathbf{b}\cdot\mathbf{x})\mathbf{x}-\Vert \mathbf{x}\Vert^2\mathbf{b},\\
    \dot{\mathbf{y}}(t)&= \frac{\dot{\mathbf{x}}}{\Vert\mathbf{x}\Vert^2}-2\frac{\mathbf{x}\cdot\dot{\mathbf{x}}}{\Vert\mathbf{x}\Vert^4}\mathbf{x}=-\mathbf{b}.
\end{align}
We now get the incredibly simple solution $\mathbf{y}(t)=\mathbf{y}_0-t\mathbf{b}$, a translation, after which we can undo the inversion
\begin{equation}
    \frac{\mathbf{x}(t)}{\Vert\mathbf{x}\Vert^2}= \frac{\mathbf{x}_0}{\Vert\mathbf{x}_0\Vert^2}-t\mathbf{b}.
\end{equation}
This form is equivalent to a Special Conformal Transformation (SCT) \cite{francesco2012}, which we can see by defining the finite transformation as $\mathbf{f}(\mathbf{x}_0) = \mathbf{x}(1)$, and taking the inner product of both sides with themselves
\begin{equation}
    \Vert\mathbf{f}(\mathbf{x}_0)\Vert^2= \frac{\Vert\mathbf{x}_0\Vert^2}{1-2\mathbf{b}\cdot\mathbf{x}_0+\Vert \mathbf{b}\Vert^2\Vert\mathbf{x}_0\Vert^2},
\end{equation}
and finally isolating
\begin{align}
    \begin{aligned}
    \mathbf{f}(\mathbf{x}_0) &= \frac{\Vert\mathbf{f}(\mathbf{x}_0)\Vert^2}{\Vert\mathbf{x}_0\Vert^2}\mathbf{x}_0-\Vert\mathbf{f}(\mathbf{x}_0)\Vert^2\mathbf{b}=\frac{\mathbf{x}_0-\Vert\mathbf{x}_0\Vert^2\mathbf{b}}{1- 2\mathbf{b}\cdot\mathbf{x}_0 + \Vert{\mathbf{b}}\Vert^2\Vert{\mathbf{x}_0}\Vert^2}.
    \end{aligned}
\end{align}
An example SCT is shown in Fig. \ref{fig:conf_mappings} (d), demonstrating their non-linear nature. In the process of this derivation we have learned that SCTs can be interpreted as an inversion, followed by a translation by $-\mathbf{b}$, followed by an inversion, and the infinitesimal Eq. \eqref{eq:derivation11} is recovered when the translation is small.

By composition, the four types of finite conformal mapping we have encountered, namely translations, rotations, scalings, and SCTs, generate the conformal group - the group of transformations of Euclidean space which locally preserve angles and orientation. The infinitesimal transformations we derived directly give the corresponding elements of the Lie algebra. 

Eq. \eqref{eq:conformal_transformation_repeat} also admits non-orientation preserving solutions which are not generated by the infinitesimal approach. Composing the scalings in Eq. \eqref{eq:derivation9.5} only produces finite scalings by a positive factor, i.e. $\mathbf{f}(\mathbf{x}) = e^\lambda \mathbf{x}$. Similarly, composing infinitesimal rotations does not generate reflections - non-orientation preserving orthogonal transformations that are not connected to the identity. The conformal group can be extended by including non-orientation preserving transformations, namely inversions (Fig. \ref{fig:conf_mappings} (e)), negative scalings, and reflections as in Table \ref{tab:trans_rot_scale_sct}. All of these elements still satisfy Eq. \eqref{eq:conformal_transformation_repeat}, as do their closure under composition. By Liouville's theorem, these comprise all possible conformal mappings.

The important point for our discussion is that any conformal mapping can be built up from the simple elements in Table \ref{tab:trans_rot_scale_sct}. In other words, a neural network can learn any conformal mapping by representing a sequence of the simple elements.

\subsection{Conformal Embeddings}\label{app:conformal_embeddings}

Whereas conformal mappings have been exhaustively classified, conformal embeddings have not. While the defining equations for a conformal embedding $\mathbf{g}:\mathcal{U}\to\mathcal{X}$, namely
\begin{equation}\label{eq:app_conformal_embedding_repeat}
    \mathbf{J}^T_{\mathbf{g}}(\mathbf{u}) \mathbf{J}_{\mathbf{g}}(\mathbf{u}) = \lambda^2(\mathbf{u})\mathbf{I}_{m},
\end{equation}
appear similar to those of conformal mappings, we cannot apply the techniques from Apps. \ref{app:conformal_transformations} and \ref{app:finite-conformal} to enumerate them. Conformal embeddings do not necessarily have identical domain and codomain. As such, finite conformal embeddings can not be generated by exponentiating infinitesimals.

The lack of full characterization of conformal embeddings hints that they are a richer class of functions. For a more concrete understanding, we can study Eq. \eqref{eq:app_conformal_embedding_repeat} as a system of PDEs. This system consists of $m(m+1)/2$ independent equations (noting the symmetry of $\mathbf{J}_{\mathbf{g}}^T \mathbf{J}_{\mathbf{g}}$) to be satisfied by $n+1$ functions, namely $\bg(\bu)$ and $\lambda(\mathbf{u})$. In the typical case that $n<m(m+1)/2 - 1$, i.e. $n$ is not significantly larger than $m$, the system is overdetermined. Despite this, solutions do exist. We have already seen that the most restricted case $n=m$ of conformal mappings admits four qualitatively different classes of solutions. These remain solutions when $n>m$ simply by having $\bg$ map to a constant in the extra $n-m$ dimensions.

Intuitively, adding an extra dimension for solving the PDEs is similar to introducing a slack variable in an optimization problem. In case it is not clear that adding additional functions $\bg_i, i>m$ enlarges the class of solutions of Eq. \eqref{eq:app_conformal_embedding_repeat}, we provide a concrete example. Take the case $n=m=2$ for a fixed $\lambda(u_1, u_2)$. The system of equations that $\bg(\bu)$ must solve is
\begin{align}
\begin{aligned}
\left(\frac{\partial g_1}{\partial u_1}\right)^2
{+}\left(\frac{\partial g_2}{\partial u_1}\right)^2 &= \lambda^2(u_1, u_2),\\
\left(\frac{\partial g_1}{\partial u_2}\right)^2
{+}\left(\frac{\partial g_2}{\partial u_2}\right)^2 &= \lambda^2(u_1, u_2),\\
\frac{\partial g_1}{\partial u_1}\frac{\partial g_1}{\partial u_2}+\frac{\partial g_2}{\partial u_1}\frac{\partial g_2}{\partial u_2} &= 0.\\
\end{aligned}
\end{align}

Suppose that for the given $\lambda(u_1, u_2)$ no complete solution exists, but we do have a $\bg(\bu)$ which simultaneously solves all but the first equation. Enlarging the codomain $\cX$ with an additional dimension ($n=3$) gives an additional function $g_3(\bu)$ to work with while $\lambda(u_1, u_2)$ is unchanged. The system of equations becomes
\begin{align}
\begin{aligned}
\left(\frac{\partial g_1}{\partial u_1}\right)^2
{+}\left(\frac{\partial g_2}{\partial u_1}\right)^2
{+}\left(\frac{\partial g_3}{\partial u_1}\right)^2 &= \lambda^2(u_1, u_2),\\
\left(\frac{\partial g_1}{\partial u_2}\right)^2
{+}\left(\frac{\partial g_2}{\partial u_2}\right)^2
{+}\left(\frac{\partial g_3}{\partial u_2}\right)^2 &= \lambda^2(u_1, u_2),\\
\frac{\partial g_1}{\partial u_1}\frac{\partial g_1}{\partial u_2} +\frac{\partial g_2}{\partial u_1}\frac{\partial g_2}{\partial u_2}+\frac{\partial g_3}{\partial u_1}\frac{\partial g_3}{\partial u_2} &= 0.\\
\end{aligned}
\end{align}
Our partial solution can be worked into an actual solution by letting $g_3$ satisfy
\begin{align}
\begin{aligned}
\left(\frac{\partial g_3}{\partial u_1}\right)^2&= \lambda^2(u_1, u_2) - \left(\frac{\partial g_1}{\partial u_1}\right)^2
-\left(\frac{\partial g_2}{\partial u_1}\right)^2,
\end{aligned}
\end{align}
with all other derivatives of $g_3$ vanishing. Hence $g_3$ is constant in all directions except the $u_1$ direction so that, geometrically speaking, the $u_1$ direction is bent and warped by the embedding into the additional $x_3$ dimension.

To summarize, compared to conformal mappings, with dimension-changing conformal embeddings the number of equations in the system remains the same but the number of functions available to satisfy them increases. This allows conformal embeddings to be much more expressive than the fixed set of conformal mappings, but also prevents an explicit classification and parametrization of all conformal embeddings. 

\section{Experimental Details}\label{app:experiments}

\begin{table}[h]
\small
\caption{Network parameters for each embedding $\bg$ and low-dimensional flow $\bh$}
    \centering
    \begin{tabular}{l rrrr}
        \toprule
        \multirow{2}{*}{\textsc{Method}} & \multicolumn{4}{c}{\textsc{Dataset}} \\
        \cmidrule{2-5}
         & \textsc{Ship $m=64$} & \textsc{Ship $m=512$} & \textsc{MNIST} & \textsc{CelebA} \\
        \midrule
        \textsc{CEF $\bg$}  & 270,918  & 1,647,174 & 139,460 & 23,649 \\
        \textsc{MF $\bg$}  &  2,276,508 & 2,276,508 & 3,135,428 & 2,311,212\\
        \textsc{$\bh$}  &  16,978,432 & 49,381,376 & 21,410,816 & 418,136,600\\
        \bottomrule
    \end{tabular}
    \vskip-0.3cm
\label{tbl:params}
\end{table}

\subsection{Synthetic Spherical Distribution}\label{app:sphere-experiment}

\paragraph{Model.} The conformal embedding $\bg$ was composed of a padding layer, SCT, orthogonal transformation, translation, and scaling (see App. \ref{app:finite-conformal} for the definition of SCT). The base flow $\bh$ used two coupling layers backed by rational quadratic splines with 16 hidden units.

\paragraph{Training.}
The CEF components were trained jointly on the mixed loss function in Eq. \eqref{eq:mixed-loss-function} with an end-to-end log-likelihood term for 45 epochs. The reconstruction loss had weight 10000, and the log-likelihood had weight 10. We used a batch size of 100 and a learning rate of $5 \times 10^{-3}$ with the Adam optimizer.

\paragraph{Data.} For illustrative purposes we generated a synthetic dataset from a known distribution on a spherical surface embedded in $\mathbb{R}^3$. The sphere is a natural manifold with which to demonstrate learning a conformal embedding with a CEF, since we can analytically find suitable maps $\mathbf{g}:\mathbb{R}^2\to\mathbb{R}^3$ that embed the sphere\footnote{Technically the ``north pole" of the sphere $(0,0,1)$ is not in the range of $\bg$, which leaves a manifold $\mathbb{S}^2 \backslash \{\text{north pole}\}$ that is topologically equivalent to $\mathbb{R}^2$.} with Cartesian coordinates describing both spaces. For instance consider
\begin{equation}
    \mathbf{g}=\left(\frac{2r^2z_1}{z_1^2+z_2^2+r^2},\, \frac{2r^2z_2}{z_1^2+z_2^2+r^2},\, r\frac{z_1^2+z_2^2-r^2}{z_1^2+z_2^2+r^2}\right),
\end{equation}
where $r\in \mathbb{R}$ is a parameter. Geometrically, this embedding takes the domain manifold, viewed as the surface $x_3=0$ in $\mathbb{R}^3$, and bends it into a sphere of radius $r$ centered at the origin. Computing the Jacobian directly gives
\begin{equation}
    \mathbf{J}_{\mathbf{g}}^T \mathbf{J}_{\mathbf{g}} = \frac{4r^4}{(z_1^2+z_2^2+r^2)^2}\mathbf{I}_2,
\end{equation}
which shows that $\mathbf{g}$ is a conformal embedding (Eq. \eqref{eq:conformal_transformation}) with $\lambda(\bz)=\frac{2r^2}{z_1^2+z_2^2+r^2}$. Of course, this $\mathbf{g}$ is also known as a \emph{stereographic projection}, but here we view its codomain as all of $\mathbb{R}^3$, rather than the 2-sphere.

With this in mind it is not surprising that a CEF can learn an embedding of the sphere, but we would still like to study how a density confined to the sphere is learned. Starting with a multivariate Normal $\mathcal{N}(\boldsymbol{\mu}, \mathbf{I}_3)$ in three dimensions we drew samples and projected them radially onto the unit sphere. This yields the density given by integrating out the radial coordinate from the standard Normal distribution:
\begin{align}
\begin{aligned}
    &p_\mathcal{M}(\mathbf{\phi, \theta}) = \int_0^\infty \frac{1}{(2\pi)^{3/2}}\exp\Big\{ -\frac{1}{2}\big(r^2 - 2r\left(\cos\phi \sin\theta , \ \sin\phi \sin\theta , \ \cos\theta\right)\cdot \boldsymbol{\mu} + \Vert \boldsymbol{\mu} \Vert^2\big)\Big\} r^2 dr.
\end{aligned}
\end{align}
With the shorthand $\mathbf{t} = \left(\cos\phi \sin\theta, \ \sin\phi \sin\theta, \ \cos\theta\right)$ for the angular direction vector, the integration can be performed\begin{align}
\begin{aligned}
    &p_\mathcal{M}(\mathbf{\phi, \theta}) = \frac{1}{2^{5/2}\pi^{3/2}}e^{-\Vert \boldsymbol{\mu} \Vert^2 / 2}\Big( 2 \mathbf{t}\cdot\boldsymbol{\mu}\, +\sqrt{2\pi}\left((\mathbf{t}\cdot\boldsymbol{\mu})^2+1\right)e^{(\mathbf{t}\cdot\boldsymbol{\mu})^2/2}\left(\text{erf}\left(\mathbf{t}\cdot\boldsymbol{\mu}/\sqrt{2}\right) + 1\right) \Big).
\end{aligned}
\end{align}
This distribution is visualized in Fig. \ref{fig:sphere} for the parameter $\boldsymbol{\mu}=(-1, -1, 0)$.

\subsection{Synthetic CIFAR-10 Ship Manifolds}\label{app:cifar10}

\paragraph{Dataset.} To generate the $64$- and $512$-dimensional synthetic datasets, we sample from \textit{ship} class of the pretrained class-conditional StyleGAN2-ADA provided in PyTorch by \citet{karras2020}. To generate a sample of dimension $m$, we first randomly sample entries for all but $m$ latent dimensions, fix these, then repeatedly sample the remaining $m$ to generate the dataset. We use a training size of 20000 for $m=64$ and 50000 when $m=512$ of which we hold out a tenth of the data for validation when training. We generate an extra 10000 samples from each distribution for testing.

\paragraph{Models.} All models for each dimension $m \in \{64, 512\}$ use the same architecture for their $\bh$ components: a simple 8-layer rational-quadratic neural spline flow with 3 residual blocks per layer and 512 hidden channels each. It is applied to flattened data of dimension $m$.

The baseline's embedding $\bg$ is a rational-quadratic neural spline flow network of 3 levels, 3 steps per level, and 3 residual blocks per step with 64 hidden channels each. The output of each scale is reshaped into $8\times 8$, and the outputs of all scales are concatenated. We then apply an invertible $1 \times 1$ convolution, and project and flatten the input down to $m$ dimensions.

On the other hand, both CEFs use the same conformal architecture for $\bg$. The basic architecture follows, with input and output channels indicated in brackets. Between every layer, trainable scaling and shift operations were applied.
\begin{align*}
    \bx\ (3 \times 64 \times 64) &\rightarrow 8\times 8\ \text{ Householder Conv }(3, 192)\\
    &\rightarrow 1 \times 1\ \text{ Conditional Orthogonal Conv }(192, 192)  \\
    &\rightarrow \text{Squeeze }(192, 3072) \\
    &\rightarrow \text{Orthogonal Transformation }(3072, m) \\
    &\rightarrow \bu\ (m)
\end{align*}

\paragraph{Training.} The sequential baseline for $m=64$ required a 200-epoch manifold-warmup phase for the reconstruction loss to converge. Otherwise, for the sequential baseline and sequential CEF, $\bg$ was trained with a reconstruction loss in a 50-epoch manifold-warmup phase. We then trained $\bh$ in all cases to maximize likelihood for 1000-epochs.
The joint CEF was trained with the mixed loss function in Eq. \eqref{eq:mixed-loss-function} for 1000 epochs. All models used weights of 0.01 for the likelihood and 100000 for the reconstruction loss.

 Each model was trained on a single Tesla V100 GPU using the Adam optimizer \cite{kingma2015} with learning rate $1\times 10^{-3}$, a batch size of 512, and cosine annealing \cite{loschilov2017warm}.

\subsection{MNIST}\label{app:mnist}

\paragraph{Models.} All MNIST models use the same architecture for their $\bh$ components: a simple 8-layer rational-quadratic neural spline flow with 3 residual blocks per layer and 512 hidden channels each. It is applied to flattened data of dimension 128.

The baseline's embedding $\bg$ is a rational-quadratic neural spline flow network of 3 levels, 3 steps per level, and 3 residual blocks per step with 64 hidden channels each. The output is flattened and transformed with an LU-decomposed linear layer, then projected to 128 dimensions. 

Both CEFs use the same conformal architecture for $\bg$. The basic architecture follows, with input and output channels indicated in brackets. Between every layer, trainable scaling and shift operations were applied.
\begin{align*}
    \bx\ (3 \times 64 \times 64) &\rightarrow 8\times 8\ \text{ Householder Conv }(1, 64)\\
    &\rightarrow 1 \times 1\ \text{ Conditional Orthogonal Conv }(64, 64)  \\
    &\rightarrow \text{Squeeze }(64, 1024) \\
    &\rightarrow \text{Orthogonal Transformation }(1024, 128) \\
    &\rightarrow \bu\ (128)
\end{align*}

\paragraph{Training.} For the sequential baseline and sequential CEF, $\bg$ was trained with a reconstruction loss in a 50-epoch manifold-warmup phase, and then $\bh$ was trained to maximize likelihood for 1000-epochs.
The joint CEF was trained with the mixed loss function in Eq. \eqref{eq:mixed-loss-function} for 1000 epochs. All models used weights of 0.01 for the likelihood and 100000 for the reconstruction loss.

 Each model was trained on a single Tesla V100 GPU using the Adam optimizer \cite{kingma2015} with learning rate $1\times 10^{-3}$, a batch size of 512, and cosine annealing \cite{loschilov2017warm}.

\subsection{CelebA}\label{app:celeba}

\paragraph{Models.} All CelebA models use the same architecture for their $\bh$ components: a 4-level multi-scale rational-quadratic neural spline flow 7 steps per level, and 3 residual blocks per step with 512 hidden channels each. It takes squeezed inputs of $24\times 8 \times 8$, so we do not squeeze the input before the first level in order to accommodate an extra level. 

The baseline's embedding $\bg$ is a rational-quadratic neural spline flow network of 3 levels, 3 steps per level, and 3 residual blocks per step with 64 hidden channels each. The output of each scale is reshaped into $8\times 8$, and the outputs of all scales are concatenated. We then apply an invertible $1 \times 1$ convolution, and project the input down to 1536 dimensions. Since this network is not conformal, joint training is intractable, so it must be trained sequentially. 

On the other hand, both CEFs use the same conformal architecture for $\bg$. The basic architecture follows, with input and output channels indicated in brackets. Between every layer, trainable scaling and shift operations were applied.
\begin{align*}
    \bx\ (3 \times 64 \times 64) &\rightarrow 4\times 4\ \text{ Householder Conv }(3, 48)\\
    &\rightarrow 1 \times 1\ \text{ Conditional Orthogonal Conv }(48, 24)  \\
    &\rightarrow 2\times 2\ \text{ Householder Conv }(24, 96) \\
    &\rightarrow 1 \times 1\ \text{ Conditional Orthogonal Conv }(96, 96) \\
    &\rightarrow 1\times 1\ \text{ Householder Conv }(96, 96) \\
    &\rightarrow 1\times 1\ \text{ Orthogonal Conv }(96, 24) \\
    &\rightarrow \bu\ (24 \times 8 \times 8)
\end{align*}

\paragraph{Training.} For the sequential baseline and sequential CEF, $\bg$ was trained with a reconstruction loss in a 30-epoch manifold-warmup phase, and then $\bh$ was trained to maximize likelihood for 300-epochs.
The joint CEF was trained with the mixed loss function in Eq. \eqref{eq:mixed-loss-function} for 300 epochs. All models used weights of 0.001 for the likelihood and 10000 for the reconstruction loss.

 Each model was trained on a single Tesla V100 GPU using the Adam optimizer \cite{kingma2015} with learning rate $1\times 10^{-4}$, a batch size of 256, and cosine annealing \cite{loschilov2017warm}.
\section{Reconstructions and Samples}\label{app:ims}

\subsection{Reconstructions}\label{app:recon-ims}

\begin{figure}[h]
\centering
    \includegraphics[width=0.7\textwidth]{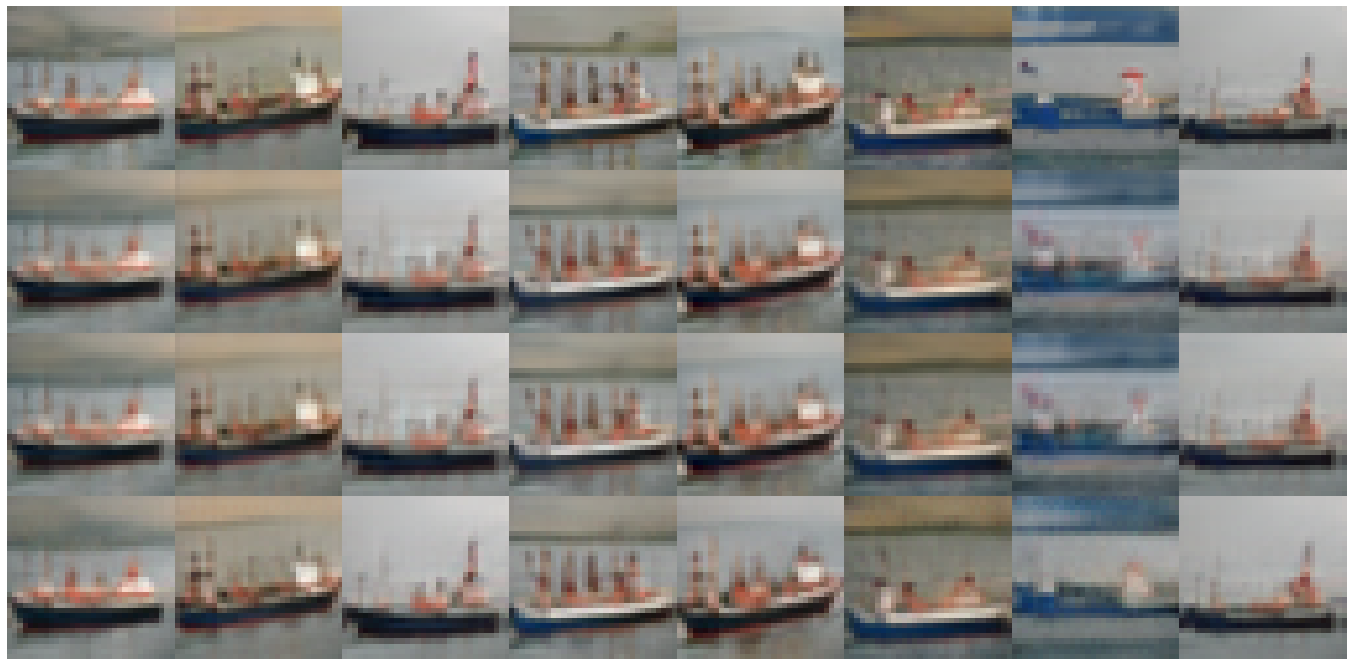}
\caption{Synthetic 64-dimensional Ship Manifold Reconstructions. From top to bottom: groundtruth samples, joint CEF, sequential CEF, and sequential MF.}
\label{fig:cifar10-64-recon}
\end{figure}

\begin{figure}[h]
\centering
    \includegraphics[width=0.7\textwidth]{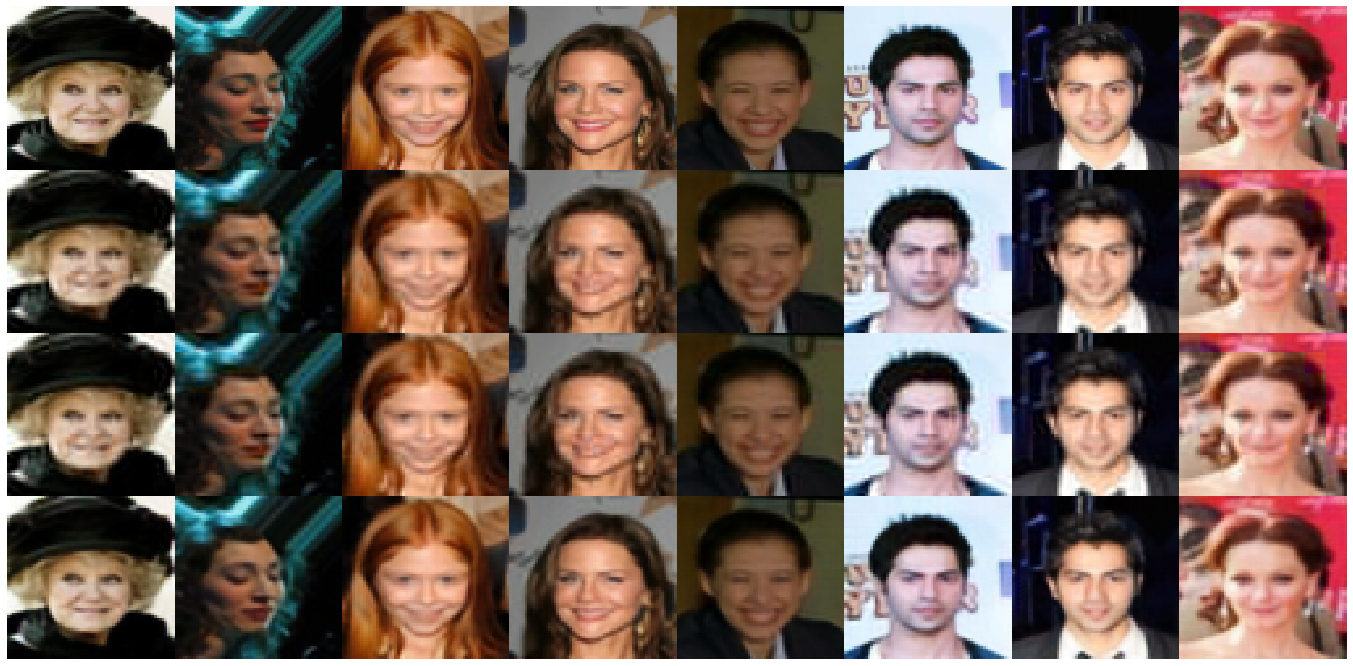}
\caption{CelebA Reconstructions. From top to bottom: joint CEF, sequential CEF, and sequential MF.}
\label{fig:celeba-recons}
\end{figure}

In this section we compare reconstructions from the remaining models omitted in the main text. These were trained on the synthetic ship manifold with 64 dimensions (Fig. \ref{fig:cifar10-64-recon}), and CelebA (Fig. \ref{fig:celeba-recons}). 

\subsection{Samples}\label{app:sample-ims}

\begin{figure}[h]
\centering
    \includegraphics[width=0.7\textwidth]{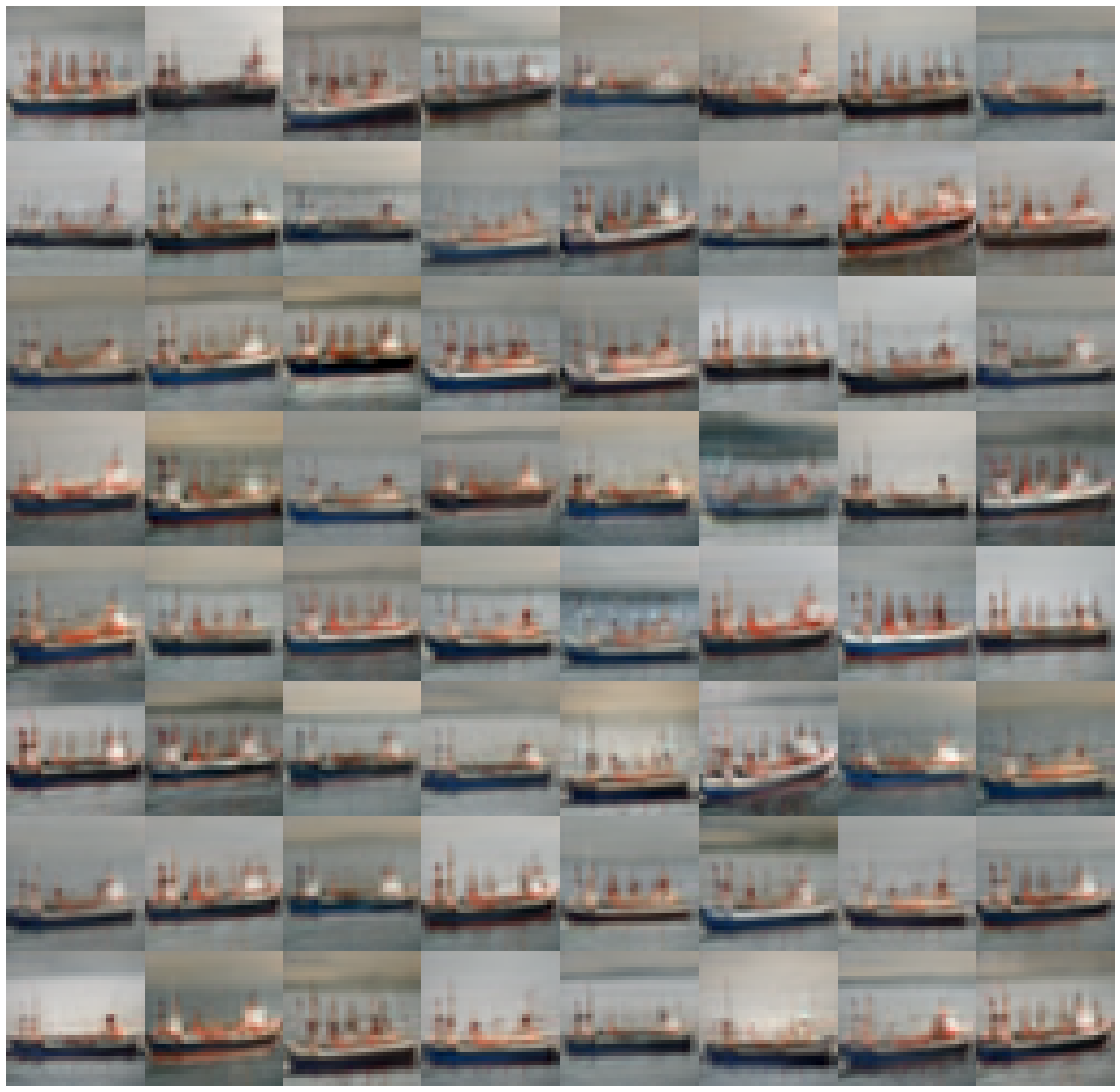}
\caption{Uncurated Synthetic 64-dimensional Ship Manifold Samples: Joint CEF}
\label{fig:ship-64-joint-cef-samples}
\end{figure}

\begin{figure}[h]
\centering
    \includegraphics[width=0.7\textwidth]{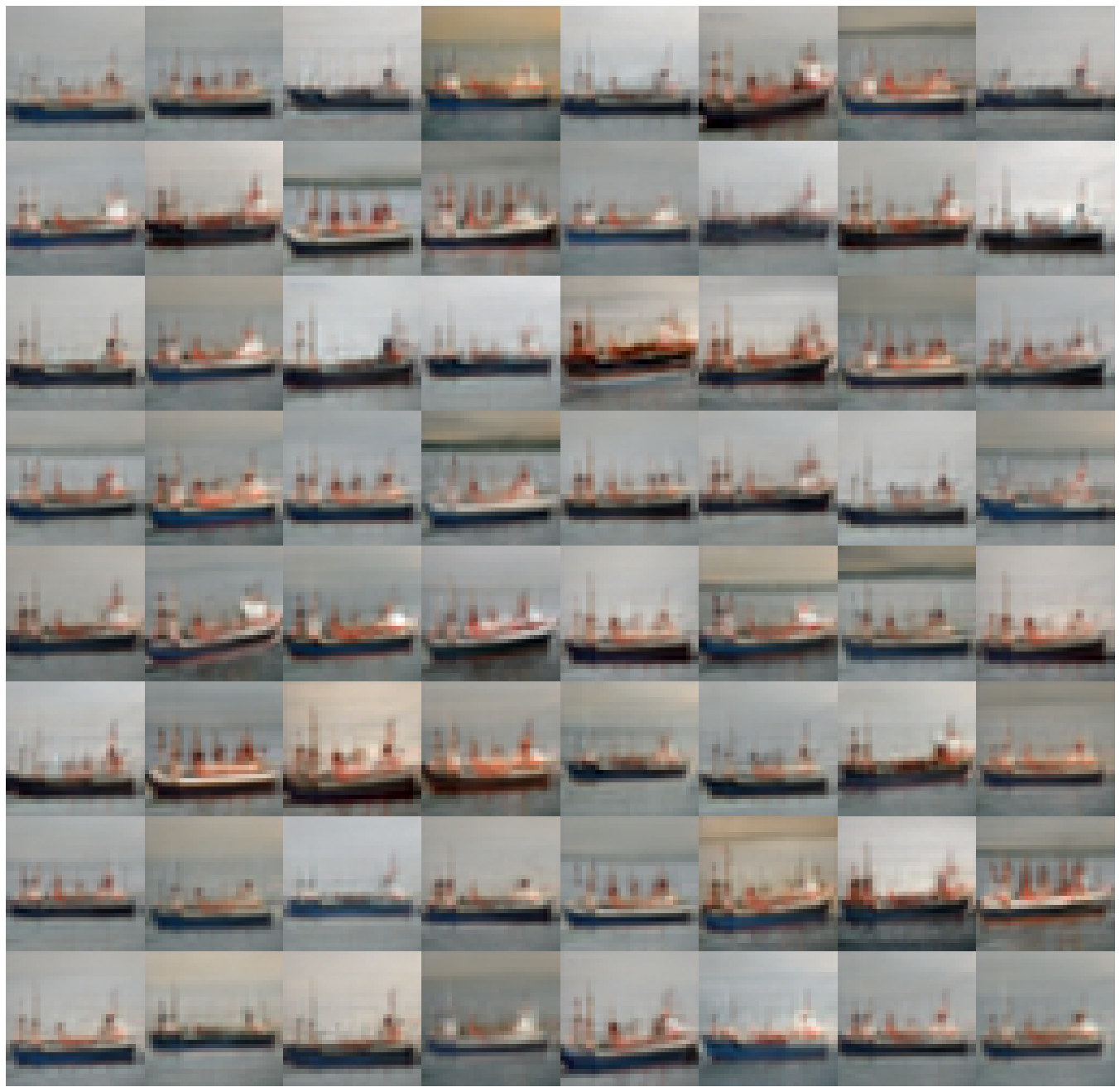}
\caption{Uncurated Synthetic 64-dimensional Ship Manifold Samples: Sequential CEF}
\label{fig:ship-64-seq-cef-samples}
\end{figure}

\begin{figure}[h]
\centering
    \includegraphics[width=0.7\textwidth]{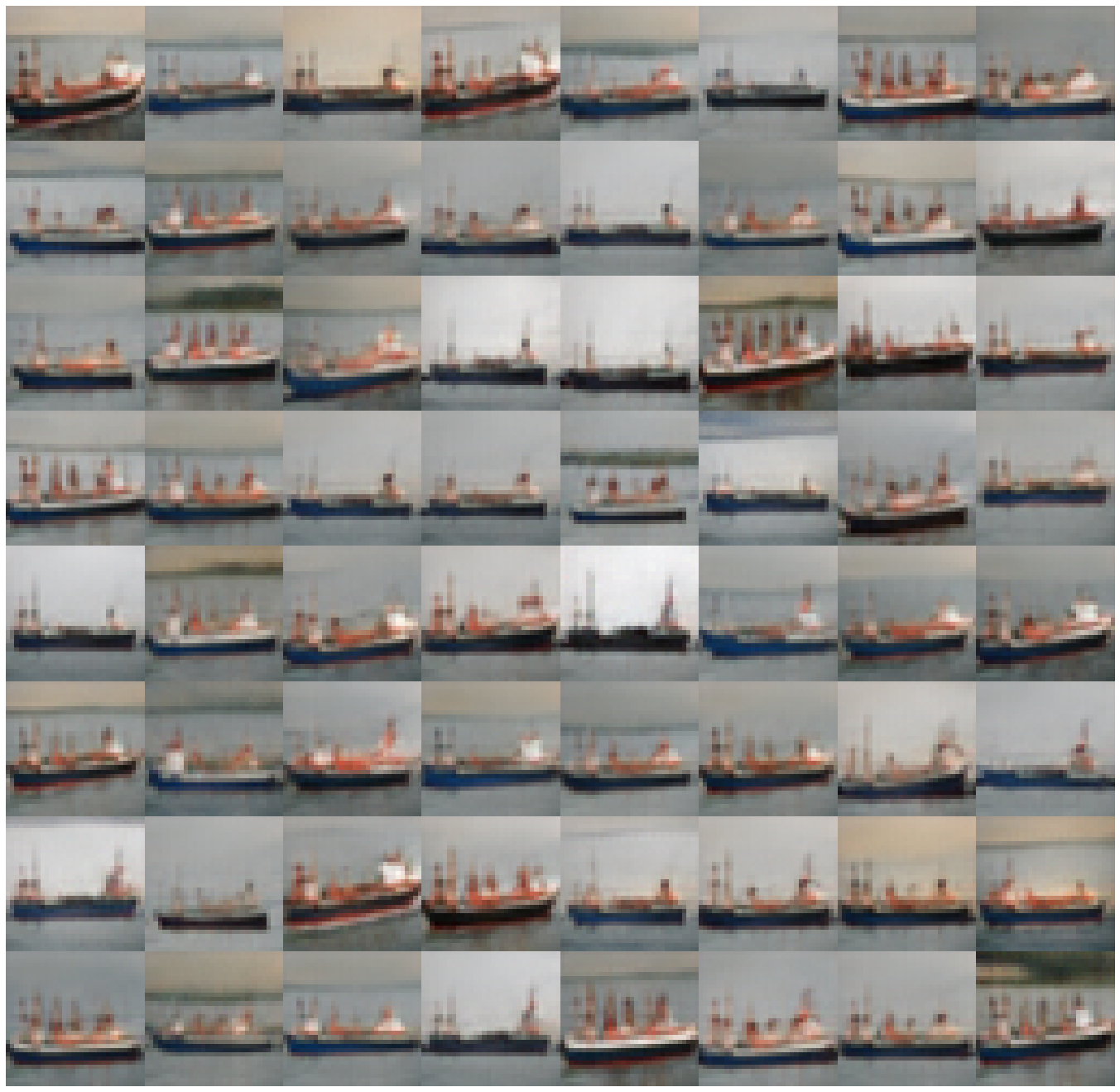}
\caption{Uncurated Synthetic 64-dimensional Ship Manifold Samples: Sequential MF}
\label{fig:ship-64-seq-mf-samples}
\end{figure}

\begin{figure}[h]
\centering
    \includegraphics[width=0.7\textwidth]{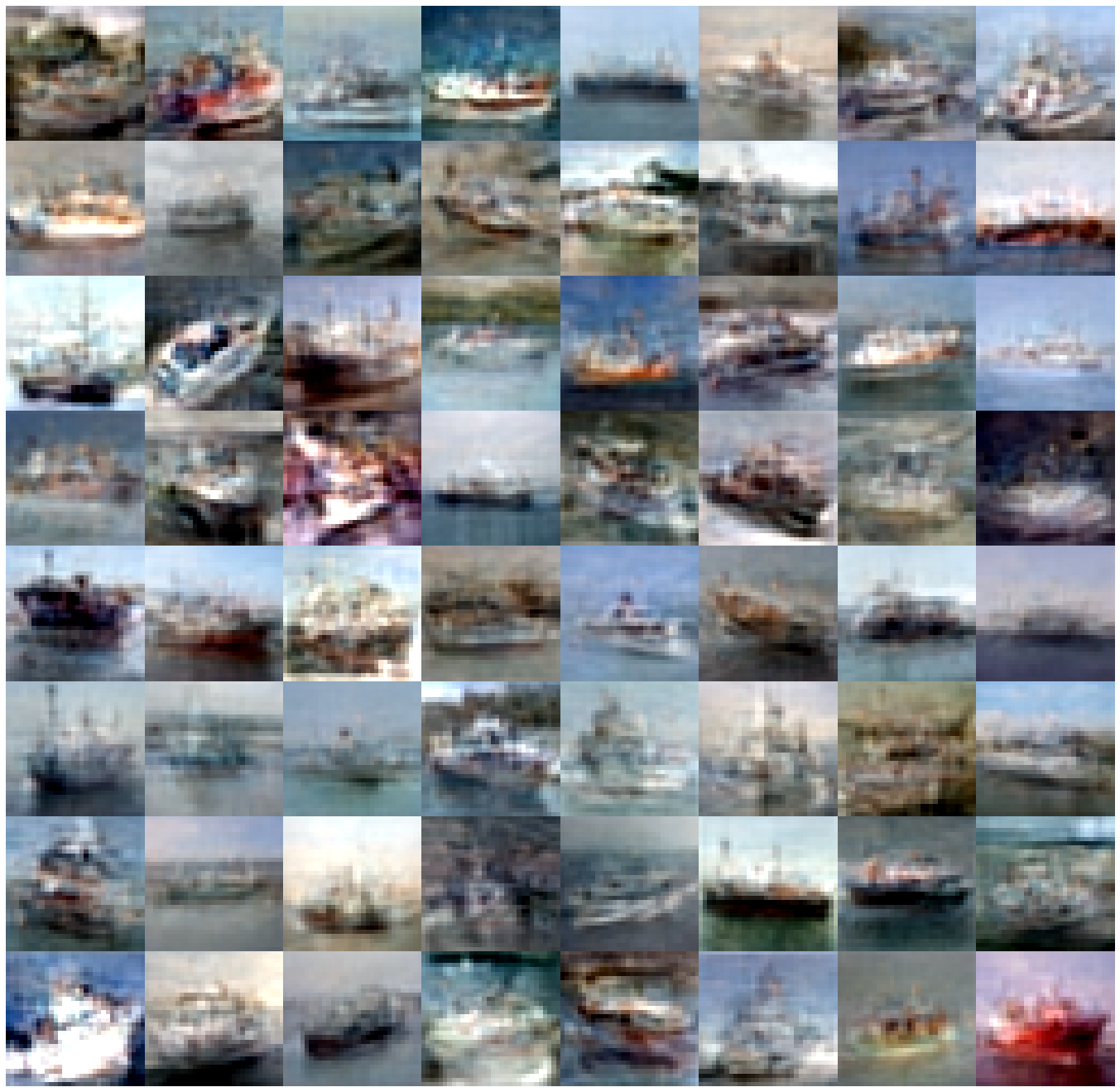}
\caption{Uncurated Synthetic 512-dimensional Ship Manifold Samples: Joint CEF}
\label{fig:ship-512-joint-cef-samples}
\end{figure}

\begin{figure}[h]
\centering
    \includegraphics[width=0.7\textwidth]{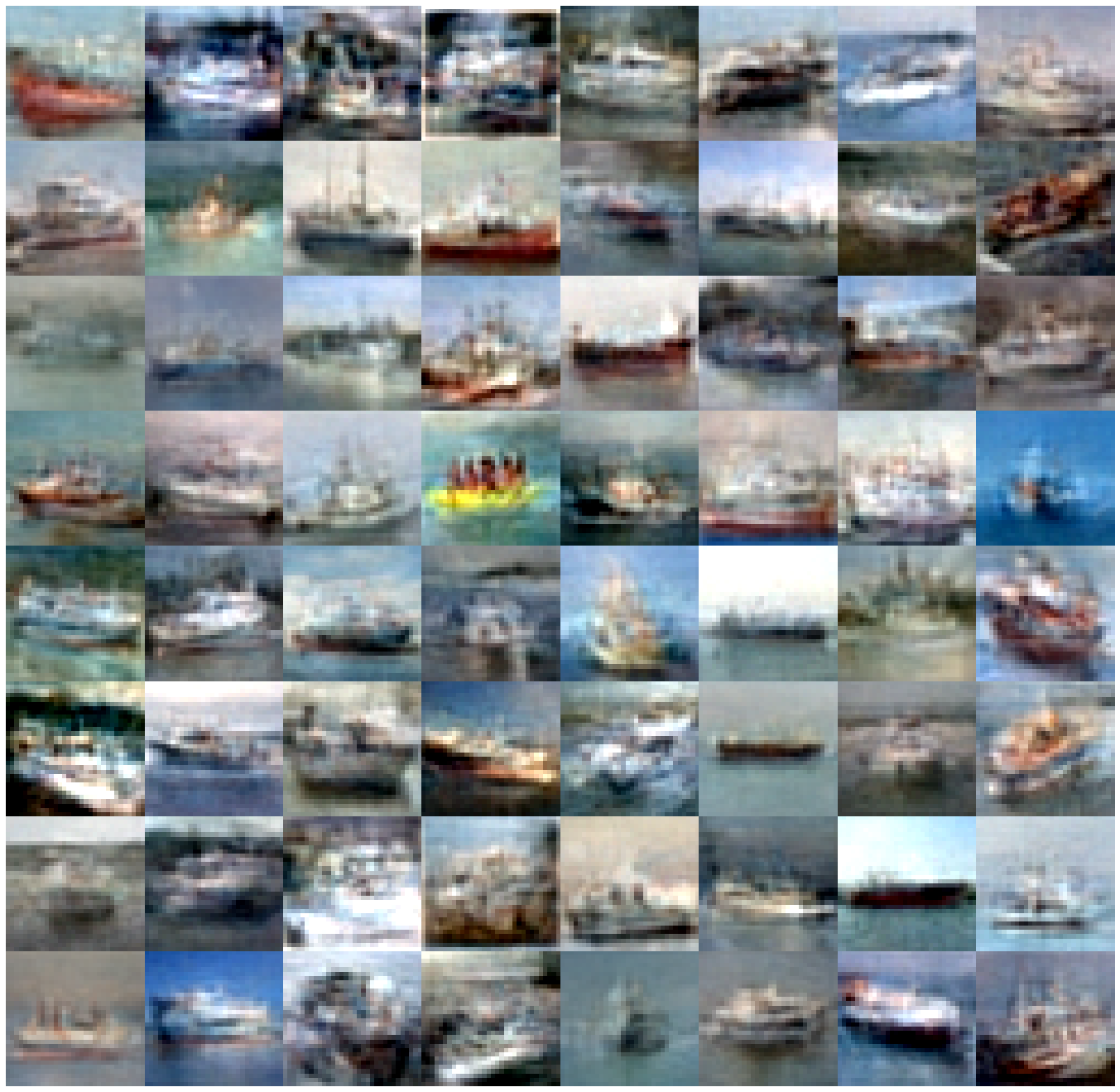}
\caption{Uncurated Synthetic 512-dimensional Ship Manifold Samples: Sequential CEF}
\label{fig:ship-512-seq-cef-samples}
\end{figure}

\begin{figure}[h]
\centering
    \includegraphics[width=0.7\textwidth]{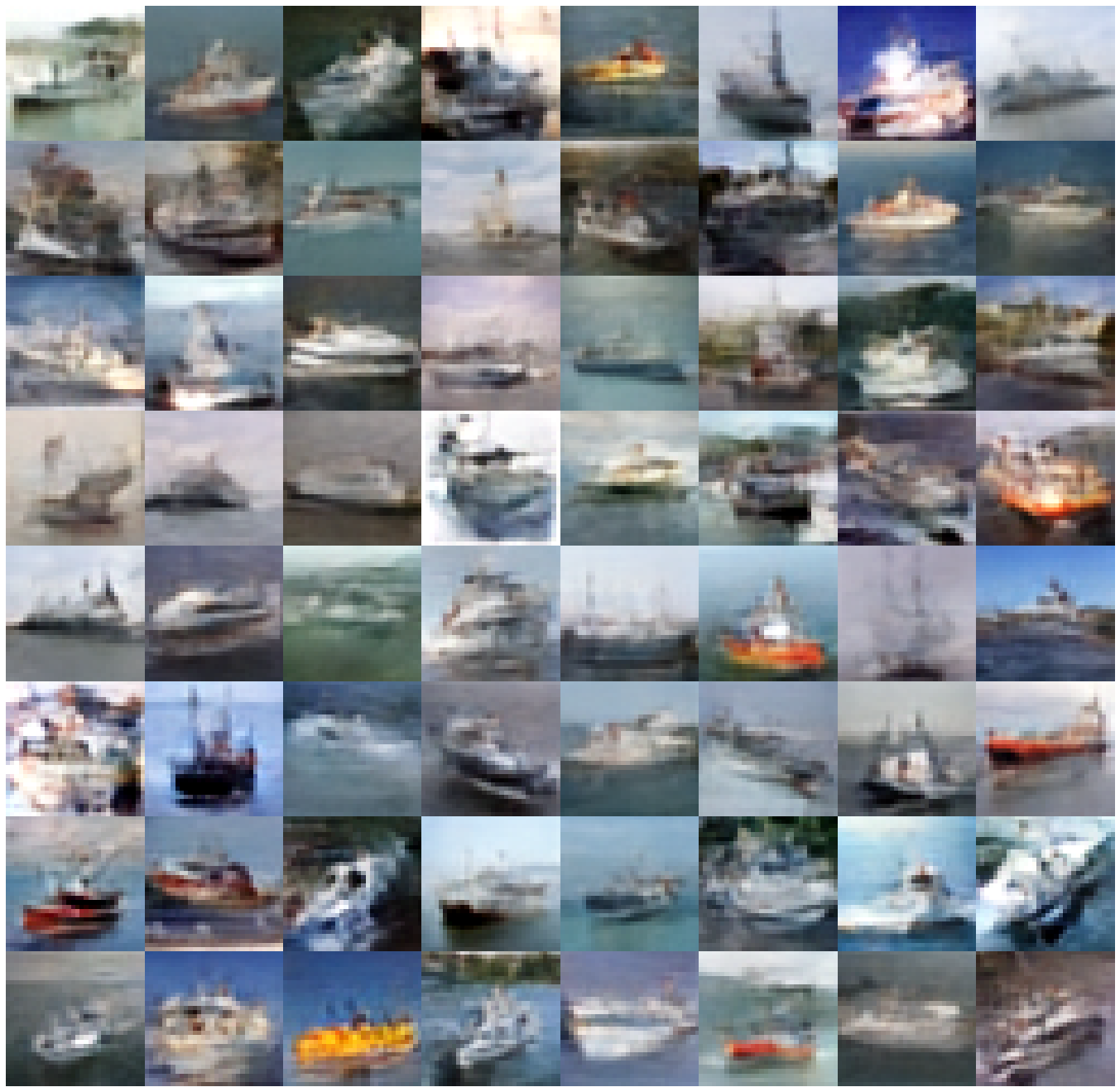}
\caption{Uncurated Synthetic 512-dimensional Ship Manifold Samples: Sequential MF}
\label{fig:ship-512-seq-mf-samples}
\end{figure}

\begin{figure}[h]
\centering
    \includegraphics[width=0.7\textwidth]{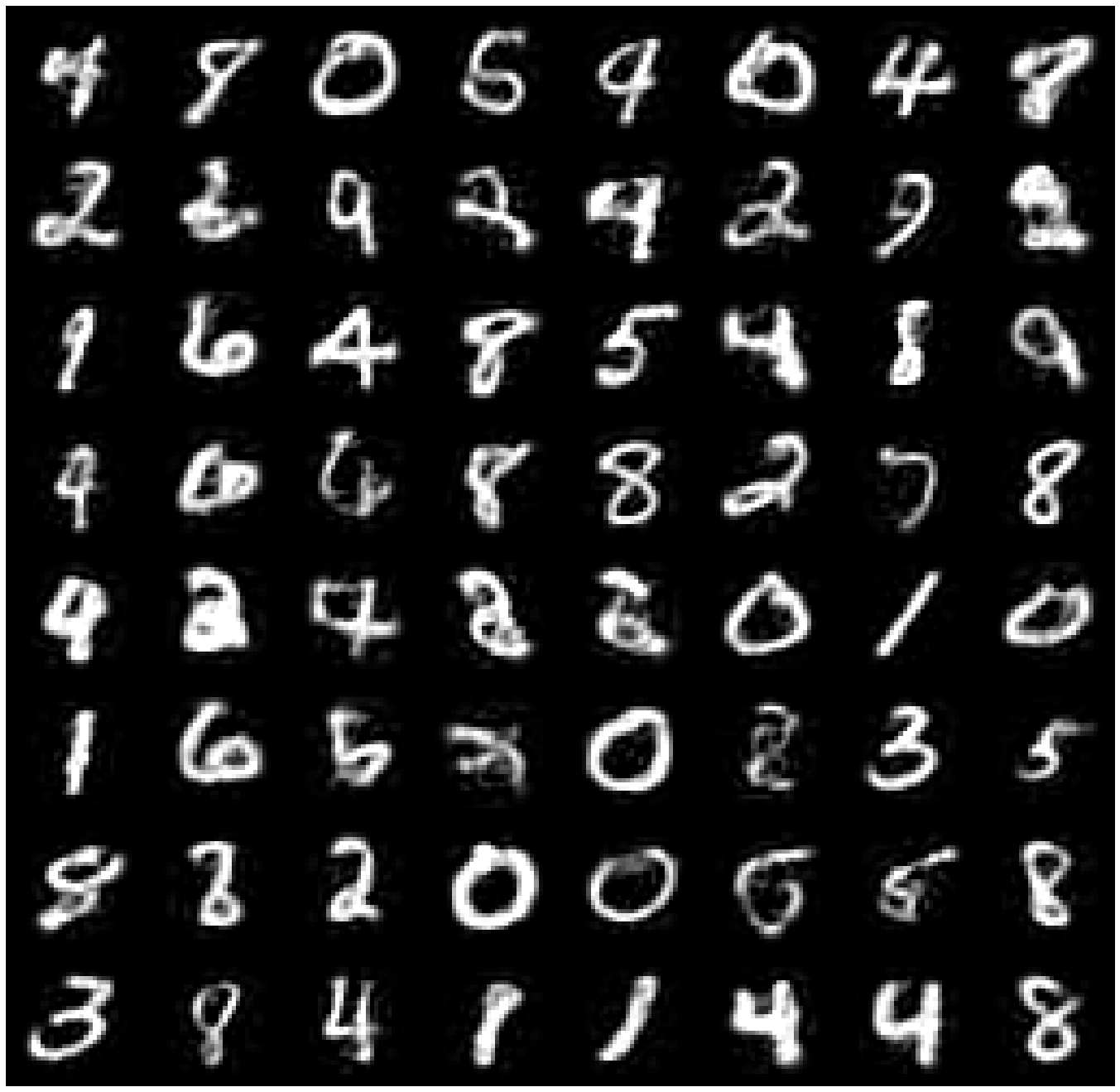}
\caption{Uncurated MNIST Samples: Joint CEF}
\label{fig:mnist-joint-cef-samples}
\end{figure}

\begin{figure}[h]
\centering
    \includegraphics[width=0.7\textwidth]{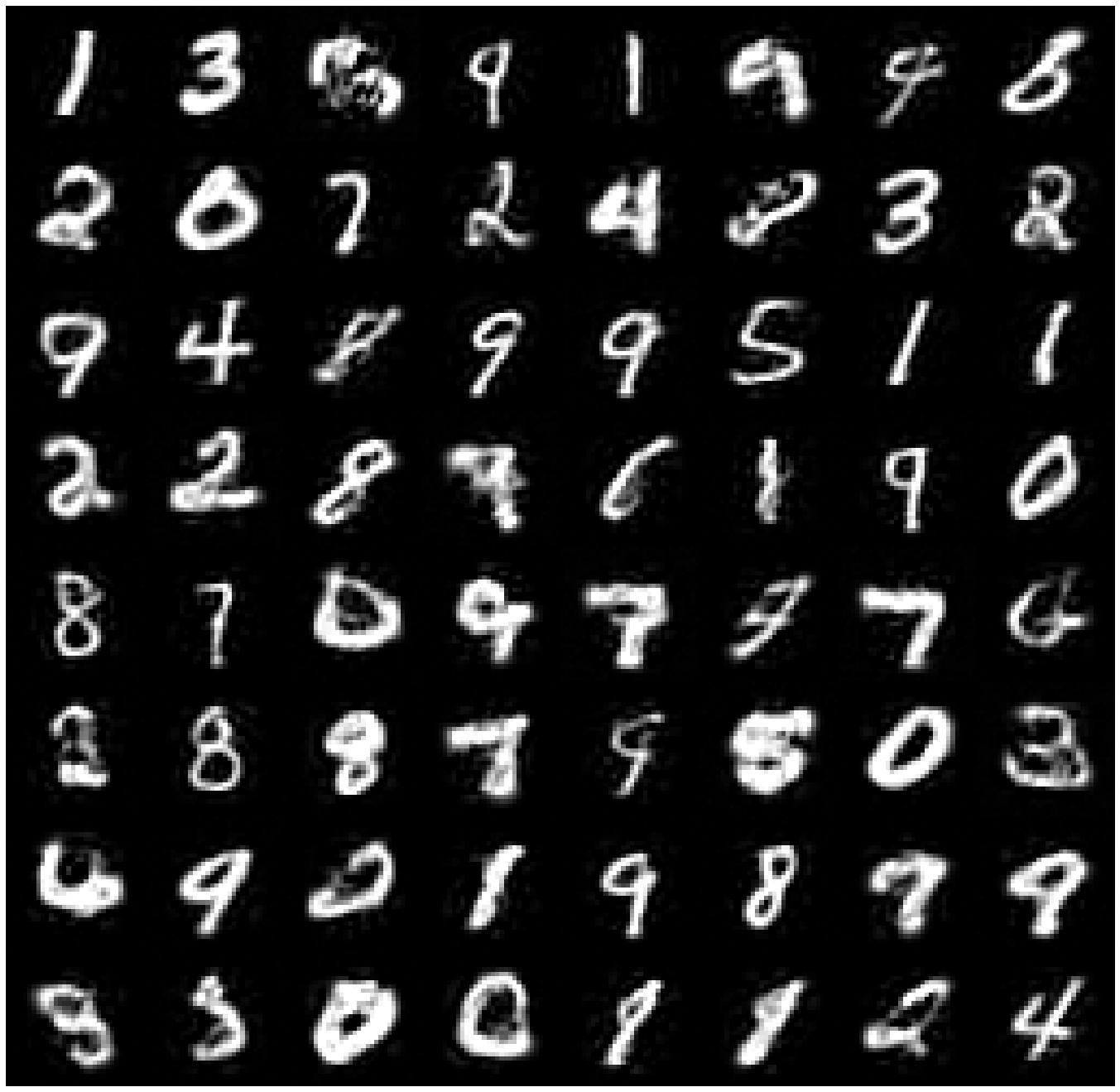}
\caption{Uncurated MNIST Samples: Sequential CEF}
\label{fig:mnist-seq-cef-samples}
\end{figure}

\begin{figure}[h]
\centering
    \includegraphics[width=0.7\textwidth]{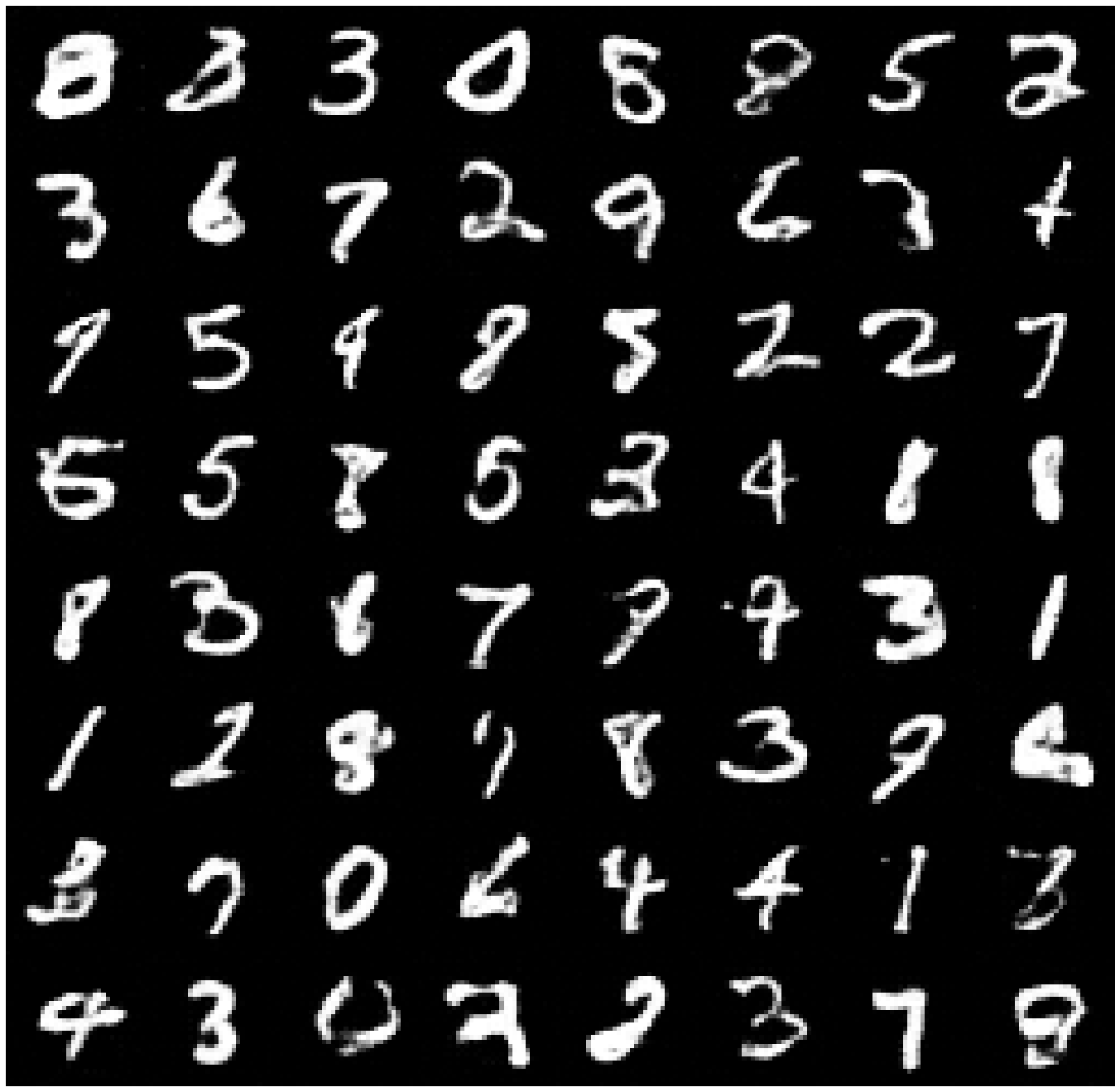}
\caption{Uncurated MNIST Samples: Sequential MF}
\label{fig:mnist-seq-mf-samples}
\end{figure}

\begin{figure}[h]
\centering
    \includegraphics[width=0.7\textwidth]{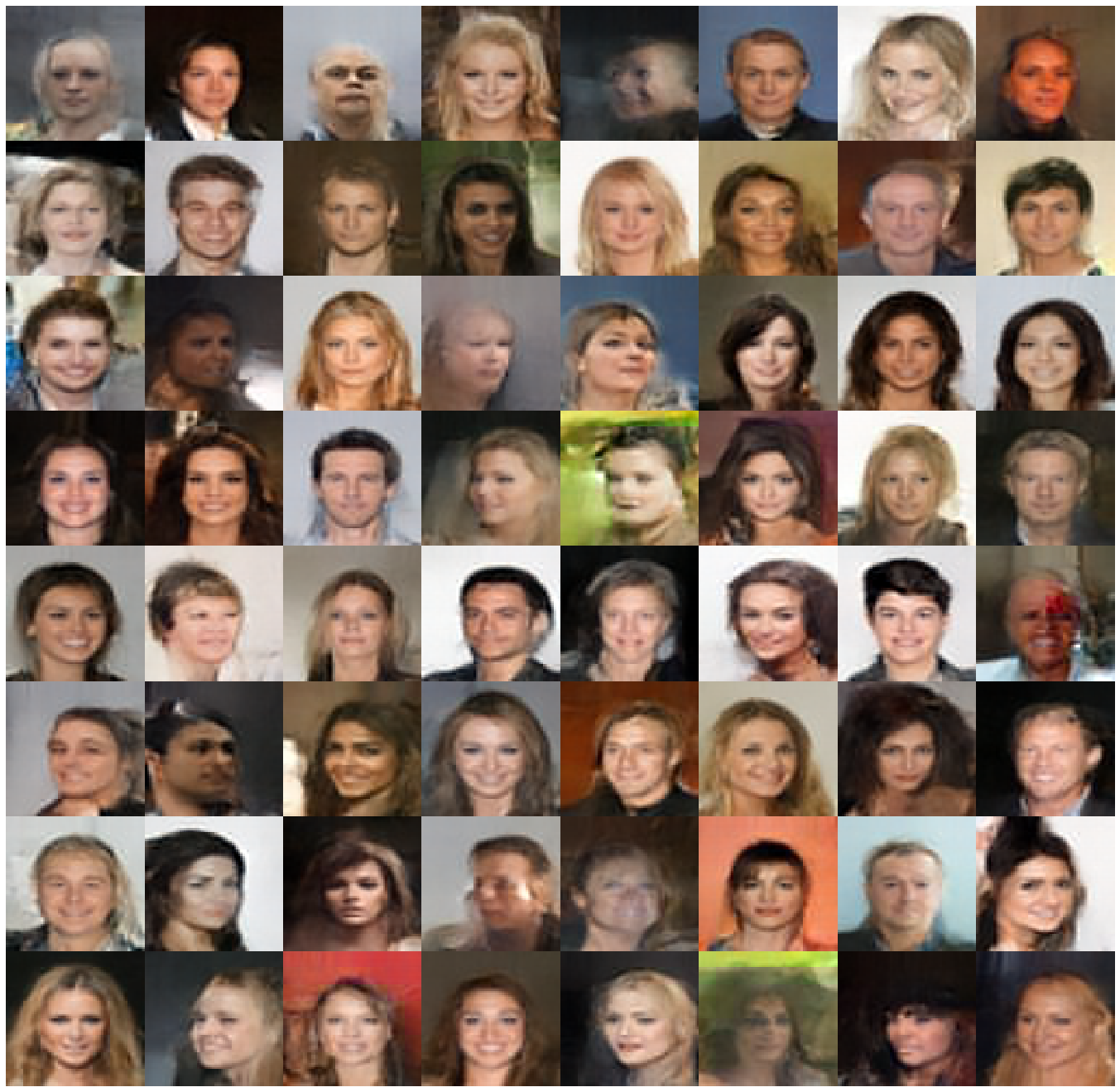}
\caption{Uncurated CelebA Samples: Joint CEF}
\label{fig:celeba-joint-cef-samples}
\end{figure}

\begin{figure}[h]
\centering
    \includegraphics[width=0.7\textwidth]{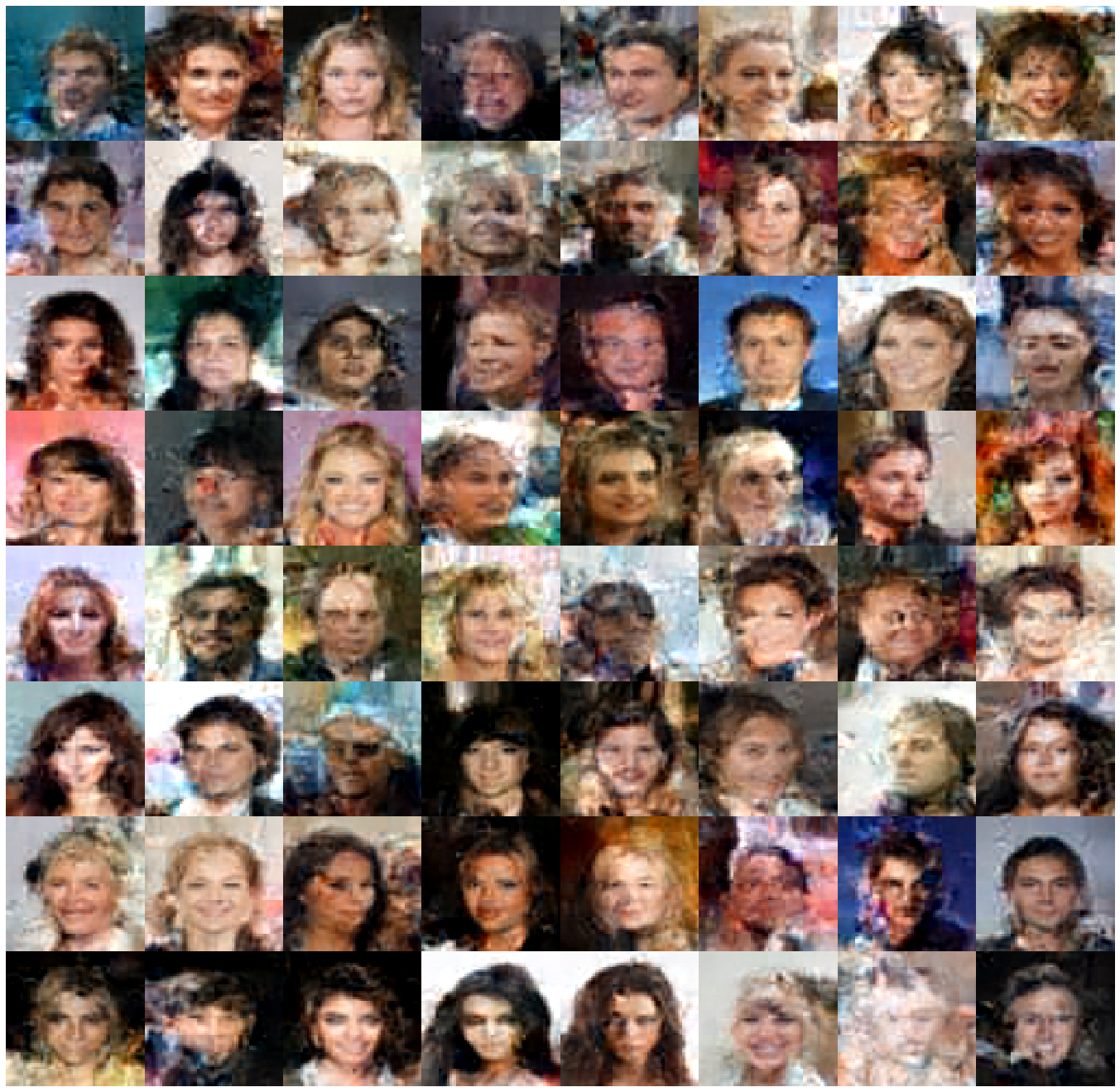}
\caption{Uncurated CelebA Samples: Sequential CEF}
\label{fig:celeba-seq-cef-samples}
\end{figure}

\begin{figure}[h]
\centering
    \includegraphics[width=0.7\textwidth]{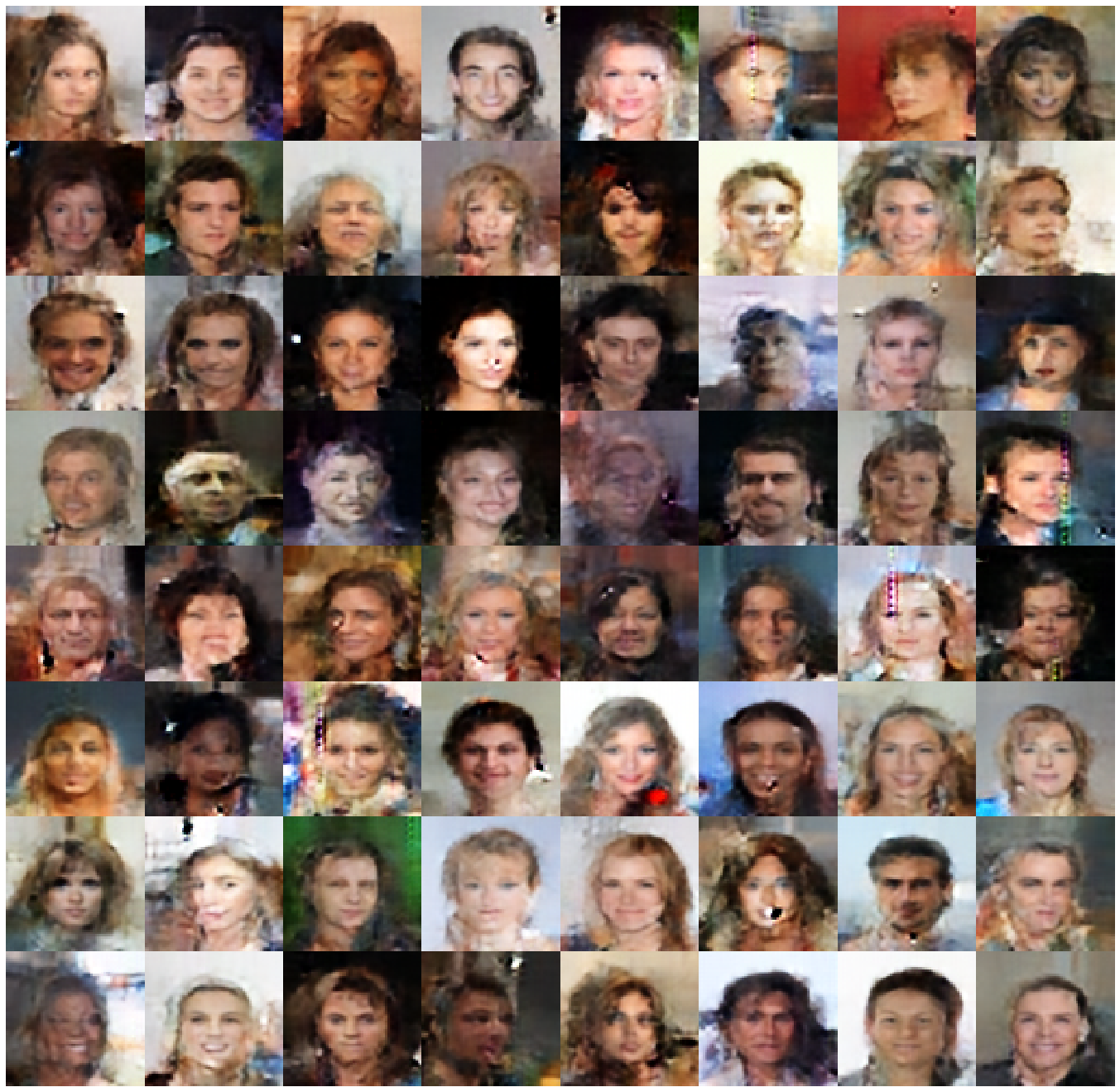}
\caption{Uncurated CelebA Samples: Sequential MF}
\label{fig:celeba-seq-mf-samples}
\end{figure}

In this section we provide additional samples from all the image-based models we trained. Figs. \ref{fig:ship-64-joint-cef-samples}-\ref{fig:ship-512-seq-mf-samples} show the synthetic ship manifolds, Figs. \ref{fig:mnist-joint-cef-samples}-\ref{fig:mnist-seq-mf-samples} show MNIST, and lastly Figs. \ref{fig:celeba-joint-cef-samples}-\ref{fig:celeba-seq-mf-samples} show CelebA.

\end{document}